# TDGT: A Tabular Data Generation Toolkit supporting adaptive GPU-accelerated Bayesian mixture models, diffusion-based models, and latent-space generative modeling

**Vasileios C. Pezoulas[1], Nikolaos S. Tachos[1], Eleni Georga[1], Kostas Marias[1], Manolis Tsiknakis[1], and Dimitrios I. Fotiadis[1]**

[1]SYNTHAINA AI, Ioannina, Greece

Corresponding author: Prof. Dimitrios I. Fotiadis, email: dimitris.fotiadis30@gmail.com.

**Abstract**

The growing demand for privacy-preserving data sharing has positioned synthetic data generation as a critical component of responsible AI workflows. Yet existing tools require programming expertise, manual hyperparameter configuration, and offer no integrated evaluation, limiting adoption among domain researchers in regulated settings. Despite notable advances in generative modeling, existing solutions often lack integration of adaptive generation strategies, multi-metric evaluation, and accessible end-to-end generators within a unified web-based toolkit. In this work, we introduce TDGT (Tabular Data Generation Toolkit), a web-based toolkit for synthetic tabular data generation and fidelity assessment. TDGT introduces the Adaptive Bayesian Mixture Synthesizer (ABMS), a novel algorithm that autonomously determines the optimal number of mixture components through iterative cluster quality optimization, eliminating the need for manual hyperparameter configuration. Building upon ABMS, we further propose VAE-ABMS, a hybrid architecture that couples Variational Autoencoder-based latent space learning with adaptive Bayesian mixture synthesis, enabling high-fidelity generation of complex, nonlinear tabular distributions. For large-scale scenarios, TDGT provides a GPU-accelerated variant of ABMS leveraging CUDA-based k-means clustering and Gaussian mixture fitting. Synthetic data fidelity is assessed through eleven statistical fidelity metrics spanning distributional divergence, structural correlation, and sample-level similarity, complemented by privacy risk indicators including k-anonymity scoring and disclosure rate estimation. The web-based toolkit supports a real-time streaming interface with interactive Plotly-based visualizations. TDGT is assessed across datasets from healthcare, socioeconomic modeling, and cybersecurity domains, demonstrating consistent generation fidelity and statistical coherence across heterogeneous feature types and data scales.

## 1. Introduction

The proliferation of AI-driven systems across diverse domains, including clinical research, financial modeling, and cybersecurity analytics has intensified the demand for high-quality tabular datasets that can be freely shared, augmented, and used for model training without compromising individual privacy. Tabular data, which are characterized by heterogeneous feature types, mixed continuous and categorical variables, and complex inter-feature dependencies, remain the dominant data modality in these domains, underpinning electronic health records, financial transaction logs, insurance databases, and network traffic repositories [1]. As the volume and sensitivity of such data continue to grow, so does the trade-off between the need for open, reproducible research and the legal and ethical obligations governing data access and subject confidentiality. Synthetic data generation has emerged as a promising solution to this challenge, offering the ability to produce high-quality datasets that preserve the distributional and structural properties of the original data while disentangling the output generated from identifiable records [2, 3]. Unlike data anonymization techniques, which operate by suppressing or perturbing existing records and are vulnerable to re-identification through auxiliary information [4], synthetic data generation constructs entirely new samples, thus providing a more robust and flexible mechanism for privacy-preserving data dissemination.

The practical significance of synthetic tabular data is particularly acute in domains where data access is structurally constrained. In healthcare, patient-level data are strictly regulated by legal frameworks like the General Data Protection Regulation (GDPR) and the Health Insurance Portability and Accountability Act (HIPAA), which impose strict limitations on data sharing across institutions and national boundaries [5]. These constraints critically impede multi-site clinical studies, the external validation of predictive models, and the development of machine learning models for rare conditions, where the aggregation of distributed cohorts is essential to achieve statistically adequate sample sizes. Synthetic data generation offers a pathway to overcome these barriers by producing surrogate datasets that retain the statistical characteristics necessary for model development and validation without exposing real patient records. Beyond healthcare, analogous constraints arise in financial services, where transaction-level data carry regulatory sensitivity and competitive confidentiality [6], and in cybersecurity, where the scarcity of labeled attack traffic datasets limits the development and benchmarking of intrusion detection systems [7]. Across these domains, synthetic tabular data have emerged as an enabling technology for collaborative research, data augmentation, algorithmic stress-testing, and the democratization of access to representative datasets in settings where data scarcity or regulatory constraints would otherwise impede scientific and technological progress.

The existing approaches for synthetic tabular data generation span across a broad methodological spectrum, ranging from classical statistical models to deep generative architectures. Methods such as multivariate normal distribution (MVND) sampling and Gaussian mixture models (GMM) offer interpretability and computational efficiency but impose restrictive distributional assumptions that inadequately capture multimodal, skewed, or heavy-tailed feature dependencies prevalent in real-world tabular data [3, 8]. The rise of deep generative modeling has substantially expanded the representational capacity of synthetic data generators beyond these parametric constraints. Generative Adversarial Network (GAN)-based approaches, exemplified by CTGAN [9] and CopulaGAN [10], employ mode-specific conditional generation strategies to handle mixed-type tabular features with heterogeneous marginals, while Variational Autoencoder (VAE)-based methods such as TVAE [9] project input data into a continuous low-dimensional latent manifold from which synthetic samples are decoded via a learned inverse mapping. Copula-based models, including GaussianCopula [10], occupy an intermediate position by decoupling the modeling of univariate marginal distributions from the estimation of inter-feature dependence structure through a copula formulation, enabling flexible joint distribution modeling without assuming multivariate normality. Despite their individual representational merits, these methods share a set of fundamental limitations that constrain their practical applicability. Key architectural choices, including the number of mixture components in GMM-based models, the dimensionality of the latent space in VAE-based architectures, and the structural priors governing copula formulations, must be specified a priori by the practitioner, introducing pronounced sensitivity to dataset-specific characteristics. This dependency on manual configuration demands considerable domain expertise and exhaustive hyperparameter optimization, both of which are rarely available in practice, particularly when generation must be applied across heterogeneous datasets spanning diverse feature types, scales, and distributional profiles. As a result, the reproducibility and transferability of these methods across data regimes remain fundamentally limited, motivating the need for adaptive, data-driven generation strategies that autonomously calibrate their internal structure to the intrinsic properties of the input data.

The practitioners who need synthetic tabular data most urgently are rarely those with the deepest generative modeling expertise. For example, a clinical data manager preparing a surrogate dataset for cross-institutional sharing, a financial analyst augmenting a small transaction log, or a cybersecurity engineer generating labeled attack traffic each faces the same practical barrier: existing tools demand hyperparameter intuition, scripting fluency, and separate evaluation pipelines that are rarely available outside of specialized research environments. Selecting the number of mixture components, tuning latent space dimensionality, or interpreting divergence metrics requires a level of generative modeling familiarity that most domain practitioners do not possess and should not need to acquire in order to benefit from synthetic data. Beyond configuration, the absence of integrated evaluation forces practitioners to assemble bespoke assessment pipelines from disparate libraries, introducing inconsistency and reproducibility concerns that undermine confidence in the generated output. The result is a persistent gap between the methodological advances of the research community and the practical accessibility of synthetic data generation in the applied settings where privacy constraints are most acute. This gap manifests differently across user profiles. For example, the journey of a domain researcher would be to upload a CSV file, select a generator, and receive a statistically evaluated synthetic dataset within minutes. An ML practitioner would need to compare a series of generators across several fidelity metrics within a single reproducible environment without integrating disparate libraries. A data infrastructure engineer would need to deploy GPU-accelerated synthesis for datasets exceeding 100,000 records without sacrificing the fidelity of the generated data.

To address these practitioners needs, we introduce TDGT (Tabular Data Generation Toolkit), an end-to-end web-based toolkit that unifies synthetic tabular data generation and multi-metric statistical evaluation within a single, accessible environment. The central methodological contribution of TDGT is the Adaptive Bayesian Mixture Synthesizer (ABMS), a novel algorithm that autonomously identifies the optimal number of Gaussian mixture components by iterating over a range of candidate cluster counts and selecting the configuration that minimizes a cluster quality criterion. This adaptive mechanism removes the dependency on manual hyperparameter specification and enables data-driven calibration of the generative model to the intrinsic structure of each dataset. Extending this principle to deep generative modeling, we further introduce VAE-ABMS, a hybrid architecture in which a Variational Autoencoder is trained to learn a compact latent representation of the input data, within which ABMS is subsequently applied to synthesize latent vectors that are decoded back to the original feature space. To address computational scalability for large-scale datasets, TDGT provides ABMS-CUDA, a GPU-accelerated implementation leveraging CUDA-based k-means clustering and Gaussian mixture fitting. Synthetic data fidelity is assessed through eleven statistical metrics including distributional divergence, structural fidelity, and sample-level similarity, complemented by privacy risk indicators. The toolkit exposes these capabilities through a real-time streaming web interface with interactive visualizations, requiring no programming expertise from the end user.

The main contributions of this work are summarized as follows:

- We propose ABMS, a novel adaptive Bayesian mixture synthesis algorithm that automatically determines the optimal number of mixture components through cluster quality optimization, enabling data-driven calibration without manual hyperparameter tuning.
- We introduce VAE-ABMS, a hybrid latent-space synthesis architecture combining Variational Autoencoder representation learning with adaptive Bayesian mixture synthesis for high-fidelity generation of complex tabular distributions.
- We present ABMS-CUDA, a GPU-accelerated variant of ABMS that enables scalable synthetic data generation for large datasets through CUDA-based clustering and mixture fitting.

- We propose TDGT, a web-based toolkit that integrates the above generation methods with an eleven-metric evaluation suite and interactive real-time visualizations, providing an end-to-end solution for synthetic tabular data generation and assessment.

The remainder of this paper is organized as follows. Section 2 presents the methods underlying each generation algorithm and the evaluation metrics implemented in TDGT. Section 3 presents experimental results across healthcare, socioeconomic, and cybersecurity datasets. Section 4 describes the TDGT web-based application technology stack. Section 5 provides a deep discussion on the findings and limitations, and Section 6 concludes the paper with directions for future work.

# 2. Methods

## 2.1. Overall pipeline

TDGT implements a modular end-to-end pipeline for synthetic tabular data generation and evaluation which is structured into four sequential stages: (i) Stage 1 - Data ingestion and feature-type detection, (ii) Stage 2 - Synthetic data generation, (iii) Stage 3 - Post-generation constraint enforcement, and (iv) Stage 4 - Multi-metric evaluation and visualization. Fig. 1 provides an overview of the full pipeline.

In the first stage, the input dataset is ingested as a comma-separated file and subjected to an automated feature-type detection procedure which classifies each column as categorical, integer-valued, or continuous floating-point, and computes per-column metadata including category registries, precision records, and empirical value bounds (Section 2.2). These metadata are propagated across all subsequent stages to ensure that the generated samples conform to the structural and representational constraints of the original data.

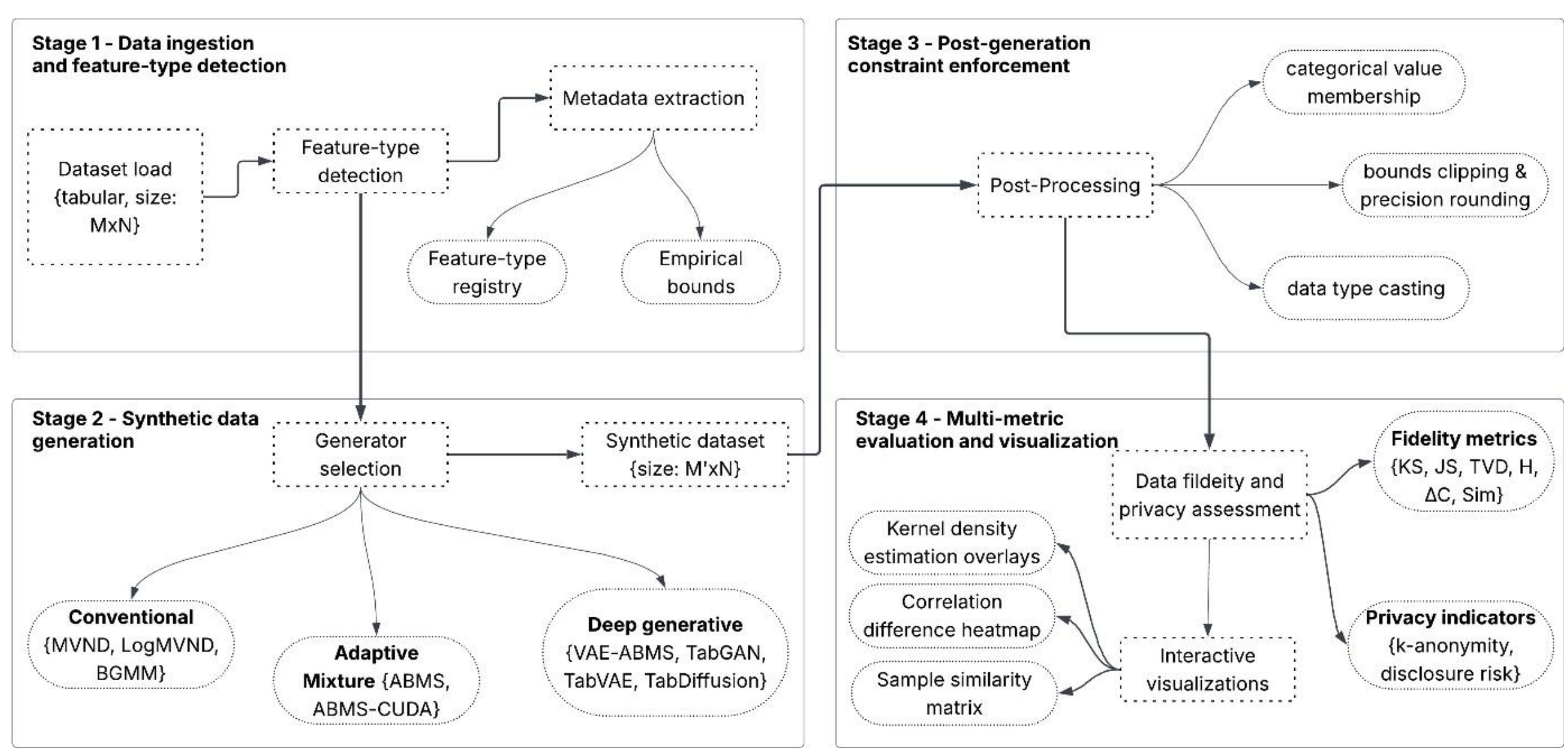

**Figure 1.** The TDGT four-stage sequential pipeline.

In the second stage, the preprocessed dataset is passed to one of nine available generators. First, three baseline statistical generators, including the Multivariate Normal Distribution sampling (MVND), the Log-transformed MVND (LogMVND), and the Bayesian Gaussian Mixture Modeling (BGMM), provide parametric reference methods of increasing expressiveness (Section 2.3.1). The Adaptive Bayesian Mixture Synthesizer (ABMS), which is introduced as one of the main methodological contribution of this work, extends the BGMM approach by autonomously identifying the optimal number of mixture components through an iterative cluster quality optimisation procedure (Section 2.3.2.1). For datasets whose complexity exceeds the representational capacity of shallow mixture models, VAE-ABMS couples a Variational Autoencoder with ABMS in the learned latent space to enable the synthesis of nonlinear and high-dimensional tabular distributions (Section 2.3.2.2). For large-scale generation scenarios, ABMS-CUDA provides a GPU-accelerated implementation of the ABMS pipeline leveraging CUDA-based clustering and mixture fitting (Section 2.3.2.3). TDGT further integrates three customised deep generative models: (i) TabGAN, a Wasserstein GAN with gradient penalty adapted for tabular synthesis (Section 2.3.3.1), (ii) TabVAE, a tabular Variational Autoencoder with adaptive batch sizing and cosine learning rate annealing (Section 2.3.3.2), and (iii) TabDiffusion, a DDPM (Denoising Diffusion Probabilistic Models)-style denoising diffusion probabilistic model with residual MLP blocks and sinusoidal time embeddings (Section 2.3.3.3).

In the third stage, all generated datasets are subjected to a uniform post-processing procedure that enforces categorical value membership, integer dtype casting, floating-point precision rounding, and empirical bounds clipping to ensure structural validity of the synthetic output regardless of the generation method employed (Section 2.4).

In the fourth stage, the synthetic dataset is evaluated against the original data through a suite of seven statistical fidelity metrics which estimate distributional divergence (KS: Kolmogorov-Smirnov statistic, KL: Kullback-Leibler divergence, JS: Jensen-Shannon divergence, TVD: Total Variation Distance, H: Hellinger distance), structural coherence (ΔC: average correlation difference), and sample-level similarity (Sim: average feature-level similarity), complemented by two privacy risk indicators (k-anonymity and disclosure risk) (Section 2.5). The evaluation results are exported as structured CSV files and rendered as interactive Plotly-based visualisations within the TDGT web interface to provide a direct visual and quantitative comparison between real and synthetic datasets.

## 2.2. Stage 1 - Data ingestion and feature-type detection

A prerequisite for accurate synthetic tabular data generation is the reliable identification of the statistical nature of each input feature, as inappropriate handling of categorical, integer, or continuous variables during generation leads to outputs that violate the structural constraints of the original data. TDGT implements an automated feature-type detection pipeline that operates directly on the input dataset prior to generation, requiring no manual annotation of column types from the user. Upon ingestion of a comma-separated input file, the dataset is loaded into a structured DataFrame and subjected to a feature classification procedure. Categorical feature detection is performed by examining the cardinality of each column, where a feature is classified as categorical if the number of distinct non-null values does not exceed a threshold of ten. For each identified categorical column, the sorted set of unique observed values is retained as a category registry, which is subsequently used during post-generation remapping to enforce valid category membership in the synthetic output. This threshold-based heuristic is consistent with established practice in tabular data modeling [11] and accommodates binary flags, ordinal scores, and low-cardinality nominal variables without requiring explicit type annotation. Numerical columns are further partitioned into integer and floating-point subtypes. A numeric column is classified as integer-valued if all non-null entries satisfy the condition that their floating-point representation is exactly integral for all observed values. Columns that do not meet this criterion are classified as continuous floating-point features. For floating-point columns, the maximum number of decimal places observed across all entries is computed from their string representations and stored as a per-column precision registry so that the synthetic values can be rounded to the original precision (post-processing).

## 2.3. Stage 2 - Synthetic data generation

### *2.3.1. Conventional generators*

TDGT provides three parametric baseline generators of increasing expressiveness: the Multivariate Normal Distribution sampler (MVND), its log-space variant (LogMVND), and a Bayesian Gaussian Mixture Model generator (BGMM).

#### 2.3.1.1. Multivariate Normal Distribution (MVND)

The simplest generator in TDGT fits a single multivariate Gaussian distribution to the input dataset. Given a real dataset $D \in R^{nxd}$ with $n$ samples and $d$ features, the empirical mean vector $\mu \in R^{d}$ and full covariance matrix $\Sigma \in R^{dxd}$ are estimated from $D$. Synthetic samples are then drawn as:

$$x_{synth} = N(\mu, \Sigma). \quad (1)$$

The full covariance structure is preserved during estimation to ensure that linear inter-feature dependencies are captured in the generated samples. While the unimodal Gaussian assumption makes MVND unsuitable for datasets with multimodal or heavily skewed marginal distributions, it provides a well-understood parametric baseline with closed-form parameter estimation and negligible computational overhead.

#### 2.3.1.2. Log-transformed Multivariate Normal Distribution (LogMVND)

Many real-world tabular datasets exhibit right-skewed or heavy-tailed marginal distributions that are poorly approximated by a Gaussian in the original feature space. LogMVND addresses this limitation by applying a log-space transformation prior to fitting. The input dataset is first transformed as:

$$\dot{D} = lo\,g(1 + D), \quad (2)$$

where $\dot{D} \in R^{\mathrm{nxd}}$ denotes the log-transformed dataset. The empirical mean and covariance are estimated from $\dot{D}$, and samples are drawn from the resulting Gaussian in log-space. The synthetic log-space samples are mapped back to the original feature space via the inverse transformation $exp(\hat{z}) - 1$, where $\hat{z} \sim N(\mu_{\dot{D}}, \Sigma_{\dot{D}})$.

#### 2.3.1.3. Bayesian Gaussian Mixture Model (BGMM)

To relax the unimodal assumptions of MVND and LogMVND, TDGT provides a Bayesian Gaussian Mixture Model generator capable of representing multimodal distributions as a weighted superposition of $K$ Gaussian components. The model is defined as:

$$p(x) = \sum_{k=1}^{K} \pi_k \, \mathcal{N}(x \mid \mu_k, \Sigma_k), \quad (3)$$

where $\pi_{\mathrm{k}}$ are the mixture weights subject to:

$$\sum_{k=1}^{K} \pi_k = 1, \ \pi_k \geq 0, \quad (4)$$

and $\mu_k, \Sigma_k$ denote the mean and covariance of the $k$-th component. Parameter estimation is performed under a Bayesian framework using variational inference, placing Dirichlet process priors over the mixture weights to allow automatic pruning of redundant components [12]. The model is fitted with $K$ components, diagonal covariance, and a regularisation term $\lambda$ added to the diagonal of each component covariance matrix to ensure numerical stability. While the Bayesian formulation permits the effective number of active components to be less than $K$ through weight concentration, the upper bound of ten components remains a fixed hyperparameter, introducing sensitivity to datasets whose intrinsic cluster structure departs significantly from this prior. This limitation directly motivates the adaptive component selection strategy introduced in Section 2.3.2.1.

### *2.3.2. Adaptive Mixture*

#### 2.3.2.1. Adaptive Bayesian Mixture Synthesizer (ABMS)

The Adaptive Bayesian Mixture Synthesizer (ABMS) extends the BGMM generator by introducing an automated cluster-count optimisation stage that determines the number of mixture components directly from the intrinsic structure of the input data, removing the dependency on manual hyperparameter specification. The algorithm proceeds in two sequential phases: (i) optimal cluster-count estimation, and (ii) adaptive Bayesian mixture fitting.

More specifically, ABMS iterates over a candidate range of cluster counts $k \in [2, K]$. At each value of $k$, $D$ is partitioned into $k$ clusters using one of four supported clustering algorithms: (i) Self-Organising Map (MiniSom, default), (ii) $k$-means, (iii) Agglomerative Hierarchical Clustering, (iv) Spectral Clustering. The quality of the partition is quantified using either the Silhouette score $S(k)$ or the Davies-Bouldin index $DB(k)$, as selected by the user. The Silhouette score measures the average ratio of intra-cluster cohesion to inter-cluster separation for each sample:

$$S(k) = \frac{1}{n}\sum_{i} \frac{b(i) - a(i)}{max(a(i), b(i))}, \tag{5}$$

where $a(i)$ is the mean intra-cluster distance of sample $i$ to all other members of its cluster, and $b(i)$ is the mean distance of sample $i$ to the nearest neighbouring cluster. $S(k) \in [-1, 1]$, with higher values indicating more compact and well-separated clusters. The Davies-Bouldin index provides a complementary measure based on the ratio of intra-cluster scatter to inter-cluster distance:

$$DB(k) = \frac{1}{k}\sum_{i=1}^{k} max_j \ \frac{\sigma_i + \sigma_j}{d(c_i, c_j)}, \tag{6}$$

where $\sigma_k$ denotes the average distance of points in cluster $k$ to their centroid $c_k$, and $d(c_k, c_j)$ is the Euclidean distance between cluster centroids. Lower DB values correspond to more compact and better-separated partitions. The optimal cluster count $k^*$ is selected as:

$$k^* = argmax_k \, S(k), \tag{7}$$

when the Silhouette score is used, or:

$$k^* = argmin_k \, DB(k), \tag{8}$$

when the Davies-Bouldin index is used.

Given the estimated optimal cluster count $k^*$, ABMS fits a BGMM with number of components equal to $k *$ and diagonal covariance structure. To penalise spurious components and encourage parsimonious mixture representations, the weight concentration prior is set adaptively through an exponential decay. This ensures that as $k^*$ increases, the Dirichlet prior exerts progressively stronger regularisation, suppressing redundant components that might otherwise capture noise rather than genuine distributional structure. A covariance regularisation term is added to the diagonal of each component covariance matrix to ensure numerical stability during variational inference.

#### 2.3.2.2. VAE-ABMS: Latent-space adaptive synthesis

VAE-ABMS extends the ABMS to the deep generative setting by coupling a Variational Autoencoder (VAE) with adaptive Bayesian mixture synthesis in the learned latent representation space. This hybrid architecture addresses the limitations of shallow mixture models for datasets with complex, nonlinear inter-feature dependencies that cannot be adequately captured by Gaussian mixtures in the original feature space. The encoder consists of two fully connected layers, followed by two parallel linear projections that output the mean vector $\mu \in \mathbb{R}^{g}$ and log-variance vector $log\sigma^2 \in \mathbb{R}^{g}$ of the approximate posterior $q(z|x) \sim N(z; \mu, \sigma^2 I)$. The latent dimension adapts to the intrinsic dimensionality of the input. The decoder mirrors the encoder architecture, mapping latent vectors back to the standardised feature space. Synthetic latent samples are drawn using the reparameterisation trick:

$$z = \mu + \sigma \odot \varepsilon, \;\; \varepsilon \sim \mathcal{N}(0, I), \tag{9}$$

where the standard deviation vector $\sigma$ is obtained from the log-variance output as $exp(1/2 \; log(\sigma^2))$, $\odot$ denotes element-wise multiplication (Hadamard product) and $\varepsilon$ is a noise vector sampled from a standard normal distribution with zero mean and identity covariance. The model is trained for a fixed number of epochs with the Adam optimiser using a composite objective that combines mean squared reconstruction error with a $\beta$-weighted KL divergence penalty $D_{KL}(.)$:

$$\mathcal{L} = \mathbb{E}(\|x - \hat{x}\|^2) - \beta \, D_{KL}(q(z|x) \parallel p(z)), \tag{10}$$

where $\beta$ is the KL weight, and is adjusted to apply light regularisation pressure on the posterior, preserving reconstruction fidelity at the cost of a moderately structured latent space, and $p(z) \sim N(0, I)$. The primary goal of the VAE in VAE-ABMS is faithful reconstruction rather than disentanglement, as the subsequent mixture model is responsible for capturing the distributional structure of the encoded representations. After training, the full dataset is encoded into the latent space by sampling the posterior using *eq. (9)*, yielding a latent representation matrix $Z \in \mathbb{R}^{nxl}$ where $l$ is the latent dimension. The latent representations are standardised via a whitening transformation, yielding $\tilde{Z}$ to ensure numerical stability of the subsequent Bayesian mixture fitting. Synthetic latent vectors are sampled from the fitted mixture, and the whitening transformation is inverted to recover samples in the original latent coordinate system. The synthetic latent vectors are finally passed through the trained VAE decoder to produce standardised synthetic feature vectors.

2.3.2.3. GPU-accelerated ABMS (ABMS-CUDA)

For large-scale datasets where the iterative cluster optimisation loop of ABMS incurs prohibitive computational cost on the CPU, TDGT provides ABMS-CUDA, a fully GPU-accelerated implementation of the ABMS pipeline. ABMS-CUDA replaces every computationally intensive component of the CPU pipeline with vectorised CUDA operations via PyTorch, while preserving identical algorithmic semantics. Cluster assignment is performed using Lloyd's algorithm accelerated entirely on the GPU. Centroids are initialised using the k-means++ strategy, where the first centroid is selected uniformly at random, and each subsequent centroid is sampled with probability proportional to the squared distance from the nearest existing centroid. Assignment and centroid update steps are vectorised via batched matrix operations. Both the Silhouette score and the Davies-Bouldin index are computed entirely on the GPU. The Silhouette score is evaluated using chunked Euclidean distance matrices to remain within VRAM budget, with explicit correction for the zero self-distance in the intra-cluster mean computation. The Davies-Bouldin index is computed via fully vectorised pairwise centroid distance and per-cluster scatter calculations.

The optimal cluster count $k^*$ is selected using the same criterion as the CPU ABMS (Section 2.3.2.1). A GMM with diagonal covariance is fitted on the GPU using a GPU-native GMM fitting library [13]. To address numerical underflow arising from high-dimensional diagonal Gaussian density evaluation, the input data are standardised and then projected to a maximum of $S$ principal components via truncated SVD computed on the GPU. The GMM is fitted and sampled in this reduced PCA space, and the resulting synthetic samples are mapped back to the original standardised feature space. In the event of GPU failure (e.g., insufficient VRAM or NaN sample values), ABMS-CUDA automatically falls back to the CPU implementation.

### *2.3.3. Deep generative*

#### 2.3.3.1. Customized GAN (TabGAN)

TabGAN implements a Wasserstein Generative Adversarial Network (WGAN) with Gradient Penalty (WGAN-GP) [14, 15] adapted for tabular data synthesis. The adversarial architecture consists of a generator $G$ and a critic $C$ (a discriminator without sigmoid output) trained to minimise the Wasserstein-1 distance between the real and generated data distributions. The Wasserstein objective provides theoretically grounded training stability by replacing the Jensen-Shannon divergence of standard GANs with the Earth Mover's distance, which remains well-defined and differentiable even when the real and generated distributions have disjointed support [14]. The gradient penalty constraint [15] enforces the 1-Lipschitz condition on the critic through soft penalisation of gradient norms at interpolated data points, avoiding the training instabilities and mode collapse associated with weight clipping.

The generator $G$ maps the noise vectors $z \in R^m$, where $m$ is the noise dimension, to synthetic feature vectors $\hat{x} \in R^d$ through a multi-layer feedforward network employing layer normalisation and LeakyReLU activations after each hidden layer. LayerNorm is used in place of Batch Normalisation to avoid instabilities that arise with small effective batch sizes during adversarial training [16]. The critic $C$ maps data vectors to an unbounded scalar score through a through a multi-layer network employing LeakyReLU activations and dropout regularisation after the first hidden layers. The critic and generator are trained adversarially for a fixed number of epochs. At each training iteration, the critic is updated $n^{c}$ times per single generator update, following the protocol of [15]. The critic loss $\mathcal{L}_C$ combines the Wasserstein distance estimate with the gradient penalty:

$$\mathcal{L}_C = \mathbb{E}(C(\hat{x})) - \mathbb{E}\big(C(x)\big) + \lambda\, GP. \quad (11)$$

The gradient penalty ($GP$) is computed over interpolated points $\tilde{x} = \alpha \cdot x + (1-\alpha) \cdot G(z)$, where $\alpha \sim U[0,1]$ is a random interpolation coefficient, as in:

$$GP = \mathbb{E}((\|\nabla C(\tilde{x})\|_2 - 1)^2), \quad (12)$$

with penalty weight $\lambda$. The generator minimises the negative critic score of its outputs as in:

$$\mathcal{L}_G = -\mathbb{E}(C(G(z))), \quad (13)$$

Both networks are optimised with Adam using a fixed learning rate and momentum parameters $\beta_1$, $\beta_2$ [17], with gradient norms clipping. Synthetic samples are generated by drawing noise vectors $z \sim N(0, I^m)$ and passing them through the trained generator as $\hat{x} = G(z)$.

#### 2.3.3.2. Customized VAE (TabVAE)

TabVAE implements a Variational Autoencoder (VAE) [18] adapted for tabular data generation. Unlike VAE-ABMS (Section 2.5), which fits a BGMM in the learned latent space before sampling, TabVAE samples new latent vectors directly from the standard normal prior $z \sim N(0, I)$ at inference time. This generation strategy is computationally lighter and is appropriate when stronger KL regularisation during training encourages the posterior to closely approximate the prior, rendering direct prior sampling reliable. Batch size and the number of training epochs are automatically scaled to the dataset size to balance training stability and computational cost. TabVAE employs a symmetric encoder–decoder architecture with layer normalisation and LeakyReLU activations throughout.

The encoder maps input vectors $x \in R^d$ through multiple hidden layers with dropout regularisation applied after each normalisation layer, followed by two parallel linear heads that project to the approximate posterior mean $\mu \in R^l$ and log-variance $log\sigma^2 \in R^l$. The latent dimension adapts to the input dimensionality. The decoder maps latent vectors $z \in R^l$ back to the standardised feature space through a mirrored multi-layer network. Latent samples are drawn using the reparameterisation trick in *eq. (9)*. TabVAE is optimised with Adam under a Cosine Annealing learning rate schedule over the full training horizon. The number of training epochs and batch size scale adaptively with dataset size to balance training stability and computational cost. The training objective $\mathcal{L}_V$ is the Evidence Lower BOund (ELBO):

$$\mathcal{L}_V = \mathbb{E}(\|x - \hat{x}\|^2) - \beta\, D_{KL}(q(z|x) \parallel p(z)), \tag{14}$$

where $\hat{x}$ denotes the reconstructed output of the decoder. The KL weight $\beta$ is adjusted to apply strong regularisation pressure on the approximate posterior, encouraging it to remain close to the standard normal prior $N(0, I)$. This is a deliberate design choice to ensure reliable prior sampling at generation time. Synthetic latent vectors are sampled directly from the standard normal prior $z \sim N(0, I^l)$ and decoded as $\hat{x} = Decoder(z)$.

2.3.3.3. Customized Diffusion Models (TabDiffusion)

TabDiffusion implements a Denoising Diffusion Probabilistic Model (DDPM) [19] adapted for tabular data generation. Diffusion models define a two-process framework: a fixed forward process that gradually corrupts data samples with Gaussian noise over $T$ discrete timesteps, reducing them progressively to isotropic Gaussian noise; and a learned reverse process that recovers the original data distribution by iteratively predicting and subtracting the noise added at each step, starting from a pure noise vector. Unlike adversarial and variational approaches, the training objective is a simple noise-prediction regression loss that requires neither adversarial optimisation nor closed-form posterior approximation, making training stable and scalable. The denoising network employs sinusoidal timestep embeddings and time-conditioned residual MLP blocks to predict the noise added at each diffusion step. TabDiffusion uses a linear variance schedule with $T$ diffusion timesteps, interpolating the per-step noise variance $\beta_t$ linearly from $\beta_1$ to $\beta_T$. The cumulative noise coefficient is $\bar{\alpha}_t = \prod_{s=1}^{t} (1 - \beta_s)$. The forward diffusion process corrupts a real standardised sample $x_0$ at timestep $t$ as:

$$x_t = \sqrt{\bar{\alpha}_t}\, x_0 + \sqrt{1 - \bar{\alpha}_t}\, \varepsilon, \tag{15}$$

where $\varepsilon \sim N(0, I^d)$ is the noise vector added at timestep $t$. This closed-form expression, derived from the Markov chain properties of the forward process, enables efficient training without sequential simulation: a noisy sample at any arbitrary timestep $t$ can be generated directly from $x_0$ in a single operation. The denoising network $\varepsilon_\theta(x_t, t)$ is a time-conditioned residual MLP (Multi-Layer Perceptron) consisting of three components. First, a sinusoidal time embedding module maps the integer timestep $t$ to a continuous representation $t_{emb} \in R^p$, where $p$ is the embedding dimension, using sinusoidal basis functions of geometrically spaced frequencies, followed by a linear projection with GELU activation [20]. Then, an input projection layer maps $x_t \in R^d$ to a hidden representation $h \in R^{h'}$, where $h'$ is the hidden dimension.

Afterwards, a series of residual blocks refine $h$ by adding a time-conditioned residual at each block, where each one applies a linear projection of $t_{emb}$ to the hidden dimension, adds it to $h$, and passes the result through LayerNorm and two successive linear layers with GELU activation, before adding the input $h$ as a skip connection. A final LayerNorm and linear projection to $R^d$ produce the predicted noise $\varepsilon_\theta(x_t, t)$.

The model is trained to predict the noise component $\varepsilon$ added at each timestep, minimising the mean squared error between predicted and true noise:

$$\mathcal{L}_{diff} = \mathbb{E}_{x_0,\varepsilon,t}(\|\varepsilon - \varepsilon_\theta(x_t, t)\|^2), \tag{16}$$

where $\varepsilon_\theta(x_t, t)$ denotes the denoising network parameterized by $\theta$, distinct by the noise vector $\varepsilon$. At each training step, a batch of real samples $x_0$ is drawn, timesteps t are sampled uniformly from $\{1, \ldots, T\}$, noisy samples $x_t$ are generated via the closed-form forward process, and the network predicts the added noise. The model is optimised with AdamW [21] under a Cosine Annealing learning rate schedule. The number of training epochs scales adaptively with dataset size. Gradient norms are clipped to ensure training stability. A new synthetic sample is generated by initialising $x_T \sim N(0, I^d)$ and iteratively applying the DDPM reverse denoising step for $t = T, T-1, \ldots, 1$:

$$x_{t-1} = \frac{1}{\sqrt{\alpha_t}}\left(x_t - \frac{1 - \alpha_t}{\sqrt{1 - \bar{\alpha}_t}}\,\varepsilon_\theta(x_t, t)\right) + \sigma_t\, z, \tag{17}$$

where $\alpha_t = 1 - \beta_t,\ z \sim N(0, I)$ for $t > 1, z = 0$ for $t = 1$, and the posterior variance is:

$$\sigma_t{}^2 = \frac{1 - \bar{\alpha}_{t-1}}{1 - \bar{\alpha}_t}\beta_t, \tag{18}$$

This posterior variance formula, derived under the Markov chain posterior, accounts for the remaining noise level at each step and ensures that the reverse chain samples are calibrated to the true data distribution at $x_0$.

## 2.4. Stage 3 – Post-generation constraint enforcement

Following generation, a structured post-processing step enforces type and domain consistency across all synthetic features. Synthetic values in categorical columns are remapped to the nearest observed category through an L1-distance nearest-neighbor assignment over the stored category registry, ensuring that no out-of-vocabulary values appear in the output. Integer-typed columns are explicitly cast to integer dtype, and floating-point columns are rounded to their stored precision.

Additionally, all continuous synthetic values are clipped to the empirical minimum and maximum bounds of the corresponding original feature, preventing the generation of out-of-range values that could arise from extrapolation in the tails of the fitted generative model. This combination of feature-type detection and post-generation constraint enforcement ensures that the synthetic dataset conforms to the structural and representational properties of the original data, independent of the generation algorithm employed.

## 2.5. Stage 4 – Multi-metric evaluation and visualization

TDGT evaluates the quality of generated synthetic datasets through a suite of seven per-feature statistical metrics and two dataset-level privacy indicators [3]. All distributional metrics are computed by first estimating smooth density functions over a fixed number of equally-spaced points spanning the joint range of the real and synthetic feature values, using Gaussian kernel density estimation (KDE). A small additive constant is applied to all density estimates to ensure numerical stability.

The following metrics are currently supported by the TDGT:

- The two-sample Kolmogorov-Smirnov Statistic (KS) statistic quantifies the maximum absolute difference between the empirical cumulative distribution functions (CDFs) of the real and synthetic feature distributions:

$$D_{KS} = sup_x \, |F_r(x) - F_s(x)|, \tag{19}$$

where $F_r(x)$ and $F_s(x)$ are the empirical CDFs of the real and synthetic samples respectively. $D_{KS} \in [0, 1]$, with lower values indicating greater distributional similarity. The associated p-value is reported alongside the statistic to indicate whether the null hypothesis of distributional equivalence can be rejected.

- The Kullback-Leibler Divergence (KL) divergence measures the information-theoretic dissimilarity between the real density $p(x)$ and the synthetic density $q(x)$:

$$D_{KL}(p \parallel q) = \int p(x) \, log \frac{p(x)}{q(x)} \, dx, \tag{20}$$

KL divergence is asymmetric and non-negative, equalling zero if and only if $p = q$.

- The Jensen-Shannon (JS) divergence is a symmetric and bounded variant of the KL divergence defined as:

$$JSD(P \parallel Q) = \frac{1}{2} D_{KL}(P \parallel M) + \frac{1}{2} D_{KL}(Q \parallel M), \tag{21}$$

where $M = ½(P + Q)$ is the pointwise average distribution. $JSD \in [0, 1]$ under the base-2 logarithm convention, providing a bounded measure of distributional dissimilarity that is more robust than KL to support mismatch between real and synthetic distributions.

- The Total Variation Distance (TVD) measures the maximum discrepancy between real and synthetic probability mass assignments over any measurable event:

$$TVD(p, q) = \frac{1}{2} \int |p(x) - q(x)| \, dx, \tag{22}$$

$TVD \in [0, 1]$, and equals zero when the two distributions are identical. It provides an intuitive upper bound on the probability that any statistical test can distinguish real from synthetic samples.

- The Hellinger distance (H) measures the geometric affinity between two probability distributions on the basis of the square root of their densities:

$$H(p, q) = \sqrt{1 - \int \sqrt{p(x) \cdot q(x)} \, dx}, \tag{23}$$

$H \in [0, 1]$, with $H = 0$ for identical distributions and $H = 1$ for mutually singular distributions. It is symmetric, satisfies the triangle inequality, and provides a metrically interpretable complement to the information-theoretic divergences above.

- The first Wasserstein distance ($W_1$), also known as the Earth Mover's Distance, quantifies the minimum cost of transforming the real distribution into the synthetic distribution, where cost is measured as the total probability mass multiplied by the distance transported.
- Structural fidelity is assessed by comparing the Pearson correlation matrices of the real and synthetic datasets across all valid feature pairs (excluding zero-variance columns). The global score is defined as the mean absolute difference over all upper-triangular off-diagonal entries ($\Delta C$):

$$\Delta C = \frac{1}{T} \sum_{i<j} \left| r_{ij} - r'_{ij} \right|, \tag{24}$$

where $r_{ij}$ and $r'_{ij}$ denote the Pearson correlation coefficients between features $i$ and $j$ in the real and synthetic datasets respectively, and $T = d(d-1)/2$ is the number of unique feature pairs. A per-feature variant is also reported, defined as the mean absolute row-wise deviation of each feature's correlation profile from its real counterpart.

- Per-feature exact-match similarity scores quantify the degree to which individual feature values in the synthetic dataset coincide with corresponding values in the real dataset. For each feature $f$, the score is computed as the fraction of (real sample, synthetic sample) pairs for which the feature values are identical:

$$FSS_f = \frac{1}{nu}\sum_i \sum_j \mathbb{1}(x_{i,f} = \hat{x}_{j,f}), \tag{25}$$

where $n$ and $u$ are the number of real and synthetic samples respectively. Additionally, a binary sample-wise similarity matrix $B \in \{0,1\}$^{$n \times u$} is computed, where $B[i,j] = 1$ if real sample $i$ and synthetic sample $j$ agree on at least $(d - 1)$ features, providing a visualisation of near-duplicate sample pairs.
- Two privacy risk indicators are computed when privacy evaluation is enabled.
  - The k-anonymity score ($k_{anon}$) is defined as the minimum count of identical quasi-identifier combinations observed across all rows of the synthetic dataset:

$$k_{anon} = min_t\left|\{j : J_j = J_t\}\right|, \tag{26}$$

    where $J_t$ denotes the tuple of all feature values in row $t$. Higher scores indicate greater resistance to re-identification through record linkage.
  - The disclosure risk score ($DR$) estimates the fraction of real samples for which an exact match exists in the synthetic dataset:

$$DR = \frac{1}{N}\sum_i \mathbb{1}(\exists j : \hat{x}_j = x_i), \tag{27}$$

    providing a direct measure of the re-identification vulnerability of the generated data.

# 3. Benchmarking

## 3.1. Benchmark datasets

Three benchmark datasets across the healthcare, financial, and cybersecurity domains were selected to evaluate the TDGT generators across structurally diverse data characteristics, sample sizes, and feature-type compositions. The datasets were chosen to represent a progression from a small, high-dimensional continuous-feature regime (Wisconsin Breast Cancer) through a moderate-scale mixed-type scenario (Bank Marketing) to a very large heterogeneous dataset with extreme imbalance (NSL-KDD). Table 1 summarises the key properties of each dataset.

**Table 1.** Benchmark datasets used in the experiments.

| Dataset | Domain | $n$ | Features | Feature types | Role in experiments |
|---|---|---|---|---|---|
| Wisconsin Breast Cancer | Healthcare | 569 | 30 | All continuous (float64) | Small, high-dimensional; tests generation precision on compact clinical data |
| Bank Marketing | Financial / Fintech | 41,188 | 21 | Mixed (11 categorical, 5 integer, 5 continuous) | Moderate scale, mixed types; tests categorical handling and fidelity at scale |
| NSL-KDD | Cybersecurity | 125,973 | 42 | Mixed (4 categorical, 23 integer, 15 continuous) | Very large, heterogeneous; stress-tests scalability and motivates ABMS-CUDA |

The Wisconsin Breast Cancer dataset [24] includes 569 patient records described by 30 continuous radiometric features derived from digitised fine-needle aspiration biopsy images. Features represent the mean, standard error, and "worst" (largest) value of ten nuclear morphology measurements: radius, texture, perimeter, area, smoothness, compactness, concavity, concave points, symmetry, and fractal dimension. A binary diagnosis label (malignant/benign) is included as a 31st column. All 30 feature values are continuous and lie on heterogeneous scales, making this dataset a demanding test of each generator's ability to reproduce high-dimensional continuous marginal distributions and their pairwise correlation structure without parametric assumptions.

The Bank Marketing dataset [25] includes 41,188 direct marketing contacts from a Portuguese bank, with the goal of predicting whether a client subscribed to a term deposit. The 21-column feature set is heterogeneous, including 11 categorical variables encode client demographics (job type, marital status, education), financial status (credit default, housing and personal loan flags), contact details (communication type, month, day of week), and previous campaign outcome; 5 integer-valued variables capture contact-level statistics and employment figures; and 5 continuous variables record macroeconomic indicators (consumer price index, consumer confidence index, Euribor 3-month rate, and number of employees). This mixed-type composition at moderate scale tests the full preprocessing pipeline and categorical feature handling capacity of each generator.

The NSL-KDD dataset [26] is a curated version of the KDD Cup 1999 intrusion detection benchmark, containing 125,973 network connection records across 42 features. The feature set consists of 4 categorical variables (protocol type, service, connection flag, and multi-class attack label), 23 integer-valued connection attributes (byte counts, flag indicators, login statistics), and 15 continuous statistical measures (rates and ratios derived from connection traffic). The dataset exhibits substantial class imbalance and multi-modal feature distributions, particularly in traffic-volume and byte-count features.

### 3.2. Experimental setup

Six generators are evaluated across all three datasets, including ABMS, ABMS-CUDA, VAE-ABMS, TabVAE, TabGAN, and TabDiffusion. Each generator is trained on the complete real dataset and configured to produce a synthetic dataset of the same cardinality $n$ as the original data, enabling direct one-to-one metric comparison without resampling artefacts. All generators receive the preprocessed dataset in which string-valued columns are encoded as integer category codes, and all synthetic outputs pass through the uniform post-processing pipeline described in Section 2.1 (categorical membership remapping, integer-type casting, precision rounding, and empirical bounds clipping). The three conventional generators (MVND, LogMVND, and BGMM) are excluded from the benchmark evaluation. These methods are provided in TDGT as accessible baselines for exploratory use, but their distributional assumptions are too restrictive for meaningful comparison against the adaptive and deep generative methods on the heterogeneous datasets evaluated here; their performance is therefore not reported in the quantitative results tables.

The full hyperparameter specifications for all six generators are provided in Supplementary Tables 1–6. For ABMS and ABMS-CUDA, no gradient-based training occurs; key parameters govern the cluster-count search (MiniSom, Davies-Bouldin metric, $K_{max} = 20$) and the Bayesian GMM fitting (diagonal covariance, regularisation $\lambda = 10^{-3}$). For the four deep generators, batch size and epoch count scale automatically with dataset size across three tiers: $n < 5{,}000$, $5{,}000 \leq n < 50{,}000$, and $n \geq 50{,}000$. All deep generators were executed on a NVIDIA Quadro GV100 GPU (32 GB VRAM, compute capability 7.0, PyTorch 2.4.0+cu121). ABMS-CUDA uses the same GPU for clustering and GMM fitting.

Distributional fidelity was assessed using five statistical metrics from Section 2.5 that are reported per feature and averaged: (i) the Kolmogorov-Smirnov statistic (KS), (ii) the Jensen-Shannon divergence (JS), (iii) the Total Variation Distance (TVD), (iv) the Hellinger distance (H), and (v) the mean absolute Pearson correlation matrix difference ($\Delta C$). The feature-level exact-match similarity score (Sim) is additionally reported for Wisconsin Breast Cancer only, as its $O(n^2)$ computation is prohibitive at larger dataset scales. Privacy metrics are computed under the same constraint.

### 3.3. Overall results

Table 2 reports mean per-feature fidelity metrics for each generator across the three benchmark datasets, together with the feature-level exact-match similarity score (Sim) where available. Bold entries mark the best value in each column within each dataset block. KS and JS are bounded in [0, 1]; TVD and H are computed on raw KDE grid densities and are not bounded.

**Table 2.** Mean per-feature generation fidelity metrics.

| Dataset | Generator | KS | JS | TVD | H | ΔC | Sim |
|---|---|---|---|---|---|---|---|
| Wisconsin Breast Cancer | ABMS | 0.147 | 0.174 | 109.65 | 3.960 | 0.184 | 0.0337 |
| | ABMS-CUDA | 0.145 | 0.148 | 206.93 | 4.563 | 0.182 | **0.0338** |
| | VAE-ABMS | **0.060** | 0.069 | 38.46 | 1.287 | 0.069 | 0.0324 |
| | TabVAE | 0.075 | 0.092 | 55.10 | 1.915 | 0.088 | 0.0323 |
| | TabGAN | 0.065 | **0.064** | 48.74 | 1.125 | **0.036** | 0.0325 |
| | TabDiffusion | 0.062 | 0.065 | **22.93** | **1.004** | 0.044 | 0.0324 |
| Bank Marketing | ABMS | 0.128 | 0.166 | 2.37 | 0.570 | 0.032 | N/A |
| | ABMS-CUDA | 0.122 | 0.159 | 2.72 | 0.609 | 0.032 | N/A |
| | VAE-ABMS | 0.080 | 0.079 | 0.79 | 0.243 | 0.017 | N/A |
| | TabVAE | 0.169 | 0.195 | 2.79 | 0.653 | 0.022 | N/A |
| | TabGAN | 0.070 | 0.078 | 0.74 | 0.225 | 0.017 | N/A |
| | TabDiffusion | **0.044** | **0.023** | **0.23** | **0.071** | **0.005** | N/A |
| NSL-KDD | ABMS | 0.192 | 0.146 | 10.97 | 0.971 | 0.057 | N/A |
| | ABMS-CUDA | 0.203 | 0.166 | 13.25 | 1.361 | 0.084 | N/A |
| | VAE-ABMS | 0.181 | 0.098 | 17.54 | 1.185 | 0.032 | N/A |
| | TabVAE | 0.252 | 0.162 | 12.46 | 1.428 | 0.058 | N/A |
| | TabGAN | 0.212 | 0.110 | **5.06** | **0.800** | 0.041 | N/A |
| | TabDiffusion | **0.172** | **0.079** | 8.92 | 0.962 | **0.030** | N/A |

On the Wisconsin Breast Cancer dataset, no single deep generator dominates across all four metrics. VAE-ABMS achieves the lowest KS (0.060) and competitive JS (0.069), confirming that its latent-space encoding captures the marginal distributions of the 30 radiometric features effectively. TabGAN achieves the best JS (0.064) and ΔC (0.036), reflecting strong preservation of both marginal and correlation structure. TabDiffusion achieves the lowest Hellinger distance (H = 1.004), while recording KS = 0.062 and ΔC = 0.044. All three outperform TabVAE (KS = 0.075, JS = 0.092) and the mixture-based generators ABMS (KS = 0.147, H = 3.960) and ABMS-CUDA (KS = 0.145, H = 4.563, ΔC = 0.182). ABMS-CUDA results are essentially identical to its CPU counterpart, confirming that GPU acceleration does not alter generation fidelity.

On the bank marketing dataset, TabDiffusion achieves a substantial margin of improvement over all other generators: JS = 0.023 and ΔC = 0.005, compared to the next-best JS = 0.079 (VAE-ABMS) and ΔC = 0.017 (TabGAN and VAE-ABMS, tied). TabGAN and VAE-ABMS produce comparable KS and H scores (KS ≈ 0.070–0.080, H ≈ 0.225–0.243), both well below ABMS (KS = 0.128, H = 0.570). ABMS-CUDA improves marginally over ABMS (KS = 0.122, JS = 0.159, ΔC = 0.032), while training in 4.0 s vs. 17.3 s — the clearest demonstration of GPU-acceleration benefit for quality-equivalent synthesis. TabVAE performs weakest among the deep generators (KS = 0.169, JS = 0.195).

On the NSL-KDD dataset, all generators show elevated KS relative to the smaller datasets, consistent with the increased distributional complexity. VAE-ABMS achieves the best KS (0.181) and JS (0.098), confirming the effectiveness of its latent-space encoding at large scale. TabDiffusion achieves the best ΔC (0.030) and a competitive JS (0.079). TabGAN records the lowest Hellinger distance (H = 0.800) and JS = 0.110. ABMS achieves a competitive JS (0.146) given its negligible training overhead. ABMS-CUDA produces slightly weaker results than ABMS (JS = 0.166, ΔC = 0.084 vs. JS = 0.146, ΔC = 0.057), attributable to the PCA projection to 20 principal components used during GPU GMM fitting (Section 2.3.2.3), which discards correlation information across the 42-dimensional feature space. TabVAE records the highest KS (0.252) and H (1.428). The Hellinger distances are higher than before across all generators which is attributable to the broad multi-modal distributions of traffic-volume features (src_bytes, dst_bytes, duration) in NSL-KDD.

## 3.4. A deeper look into the structural assessment of the synthetic data

Assessing the structural fidelity of synthetic tabular data requires metrics that capture not only marginal distributional similarity but also the preservation of inter-feature dependencies and higher-order relational structure across the full feature space. The following subsections present correlation difference matrices, per-entry scatter plots, and KDE overlays across all three benchmark datasets, providing a generator-level diagnostic of both structural and distributional fidelity beyond the aggregate scores reported in Table 2.

### *3.4.1. Wisconsin Breast Cancer dataset*

According to Fig. 2 ABMS and ABMS-CUDA show the strongest residual errors (ΔC = 0.184 and 0.182 respectively), concentrated along the morphological feature triples (radius–perimeter–area across mean, SE, and worst groups). TabGAN produces the closest correlation match (ΔC = 0.036), with TabDiffusion (ΔC = 0.044) and VAE-ABMS (ΔC = 0.069) following.

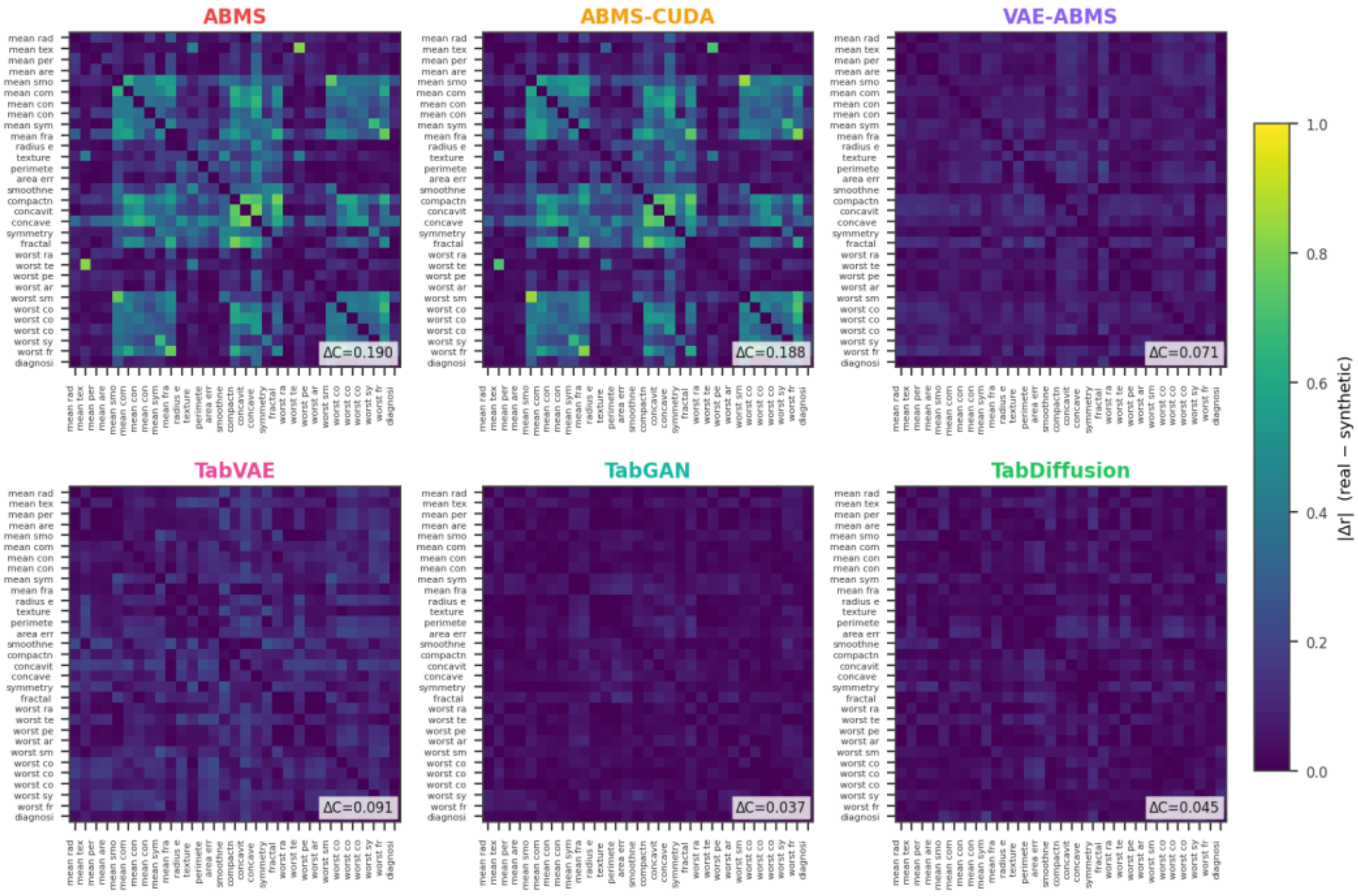

**Figure 2.** Absolute Pearson correlation difference matrices for Wisconsin Breast Cancer, one panel per generator. ΔC (mean upper-triangle error) annotated per panel.

TabGAN and VAE-ABMS exhibit the tightest scatter around the identity line (Fig. 3). ABMS and ABMS-CUDA show deviations at strong positive correlations, confirming that shallow mixture fitting does not preserve the full covariance structure of this high-dimensional continuous dataset.

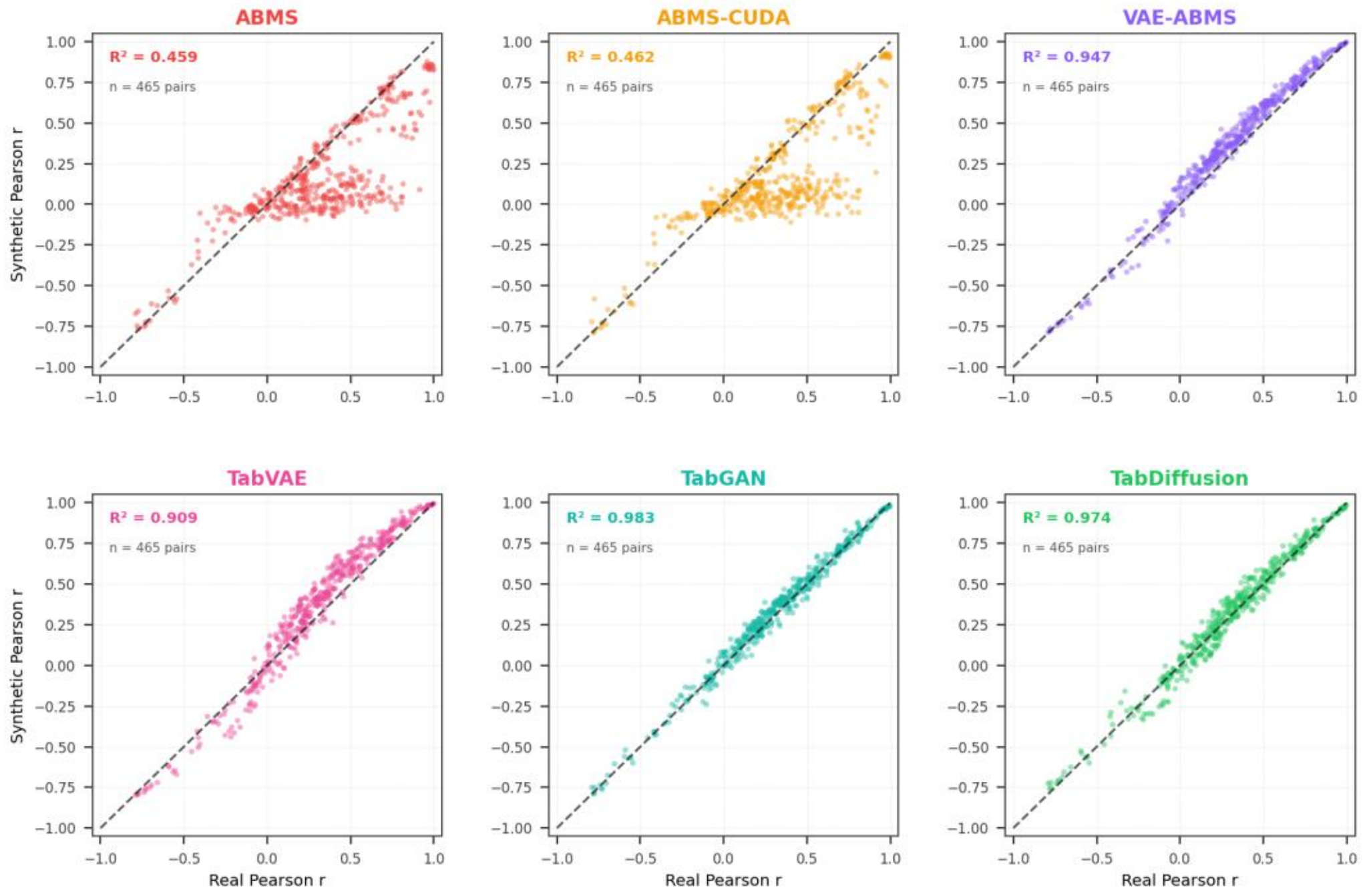

**Figure 3.** Per-entry correlation scatter plots for Wisconsin Breast Cancer. Each panel plots the real Pearson coefficient (x-axis) against the synthetic (y-axis) for each generator; dashed line indicates perfect agreement.

### *3.4.2. Bank Marketing dataset*

Fig. 4 shows the correlation difference matrices for the Bank Marketing dataset. The dominant macroeconomic indicator block (emp.var.rate, cons.price.idx, euribor3m, nr.employed) produces near-zero differences across all generators, visible as a white sub-block in the lower-right corner, confirming that all generators reproduce this strongly collinear group. Differences are larger for weakly correlated pairs between demographic and economic features and are smallest for TabDiffusion (ΔC = 0.005).

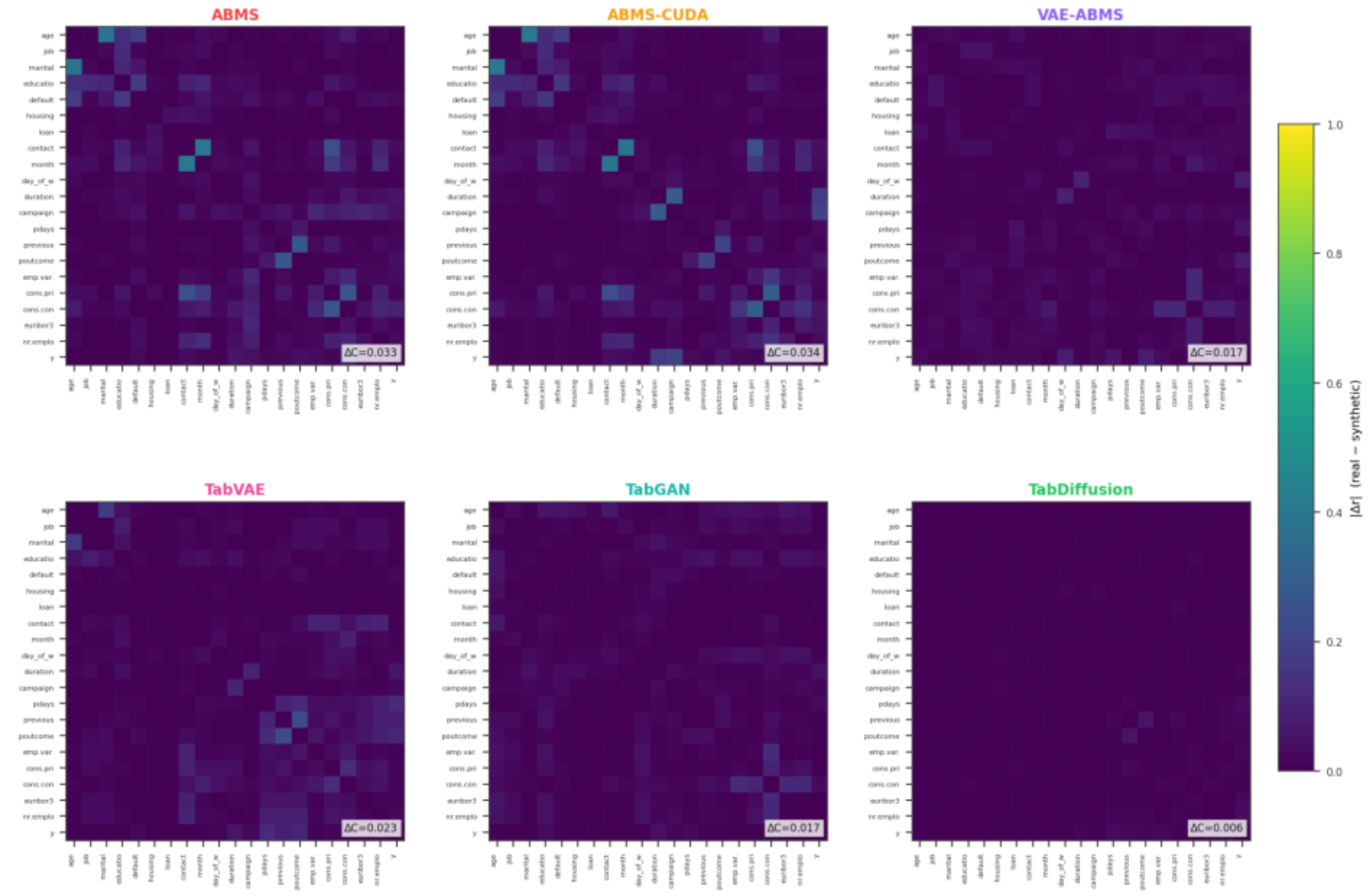

**Figure 4.** Absolute Pearson correlation difference matrices for Bank Marketing.

Fig. 5 confirms, through per-entry scatter plots, that all generators reproduce the Bank Marketing correlation structure closely. Points are tightly clustered along the diagonal for all generators, reflecting the low ΔC values reported in Table 2. TabDiffusion achieves the most concentrated scatter (ΔC = 0.005), while ABMS and TabVAE show marginally wider dispersion, primarily for weakly correlated feature pairs outside the macroeconomic block.

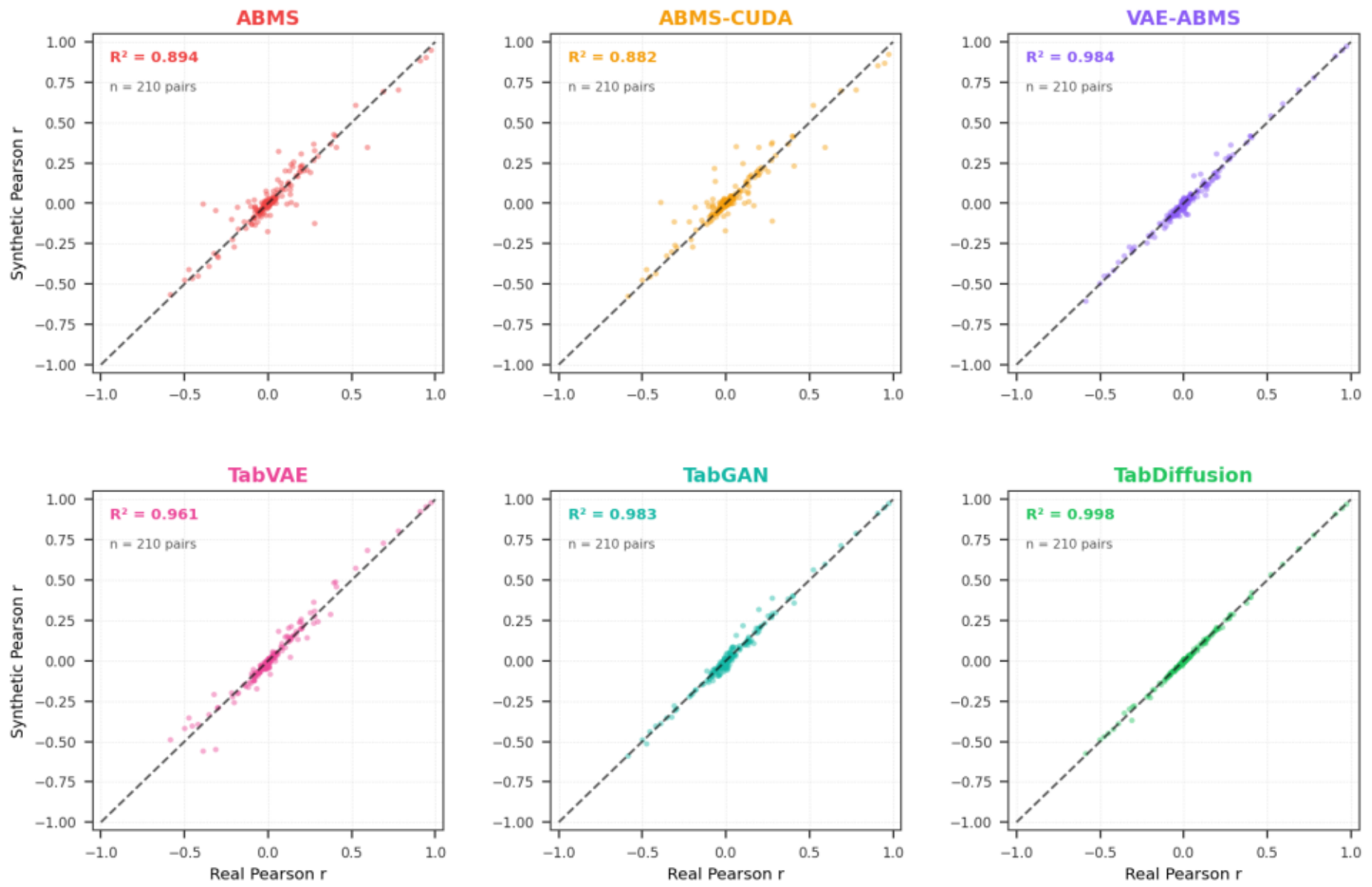

**Figure 5.** Per-entry correlation scatter plots for Bank Marketing.

### 3.4.3. *NSL-KDD dataset*

Fig. 6 shows the correlation difference matrices for NSL-KDD. The error patterns are more diffuse than on the smaller datasets, reflecting the heterogeneity of the 42-feature space. ABMS-CUDA generates the strongest residuals (ΔC = 0.084), followed by TabVAE (0.054) and ABMS (0.057); the elevated error for ABMS-CUDA reflects information loss in its PCA projection step. TabDiffusion and VAE-ABMS exhibit the most uniform low-error panels (ΔC = 0.030 and 0.032 respectively, Table 2).

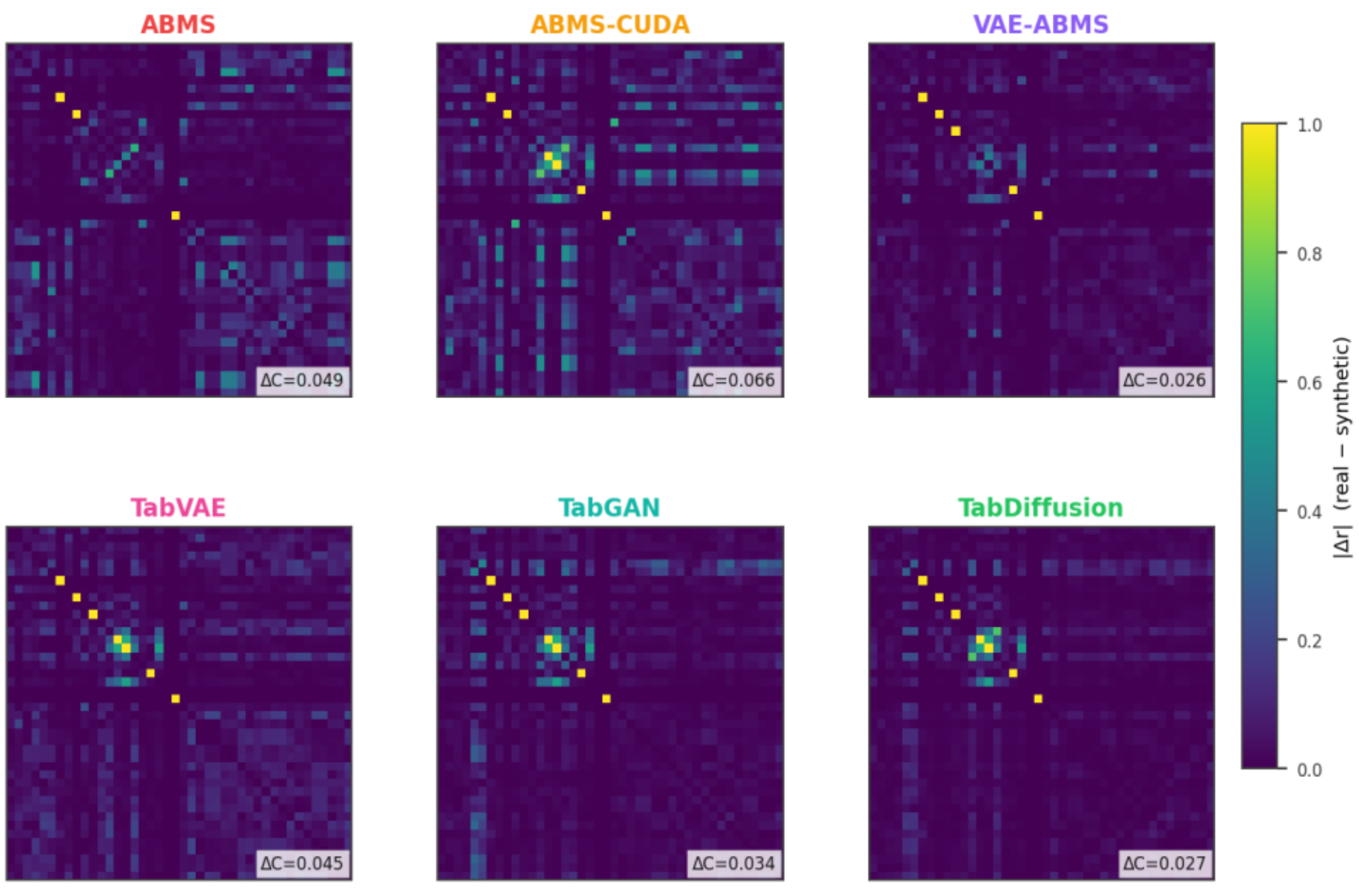

**Figure 6.** Absolute Pearson correlation difference matrices for NSL-KDD.

Fig. 7 provides the per-entry scatter plots for NSL-KDD. The wider scatter compared to the smaller-scale datasets reflects the inherent difficulty of reproducing the correlation structure of 42 heterogeneous features spanning categorical, integer, and continuous types. VAE-ABMS and TabDiffusion achieve the tightest alignment (ΔC = 0.032 and 0.030 respectively), with points closely following the diagonal even at high positive correlation values. ABMS and ABMS-CUDA both show greater spread (ΔC = 0.057 and 0.084), particularly in the moderate positive correlation range (r ≈ 0.3–0.6), where the mixture model underestimates dependency strength.

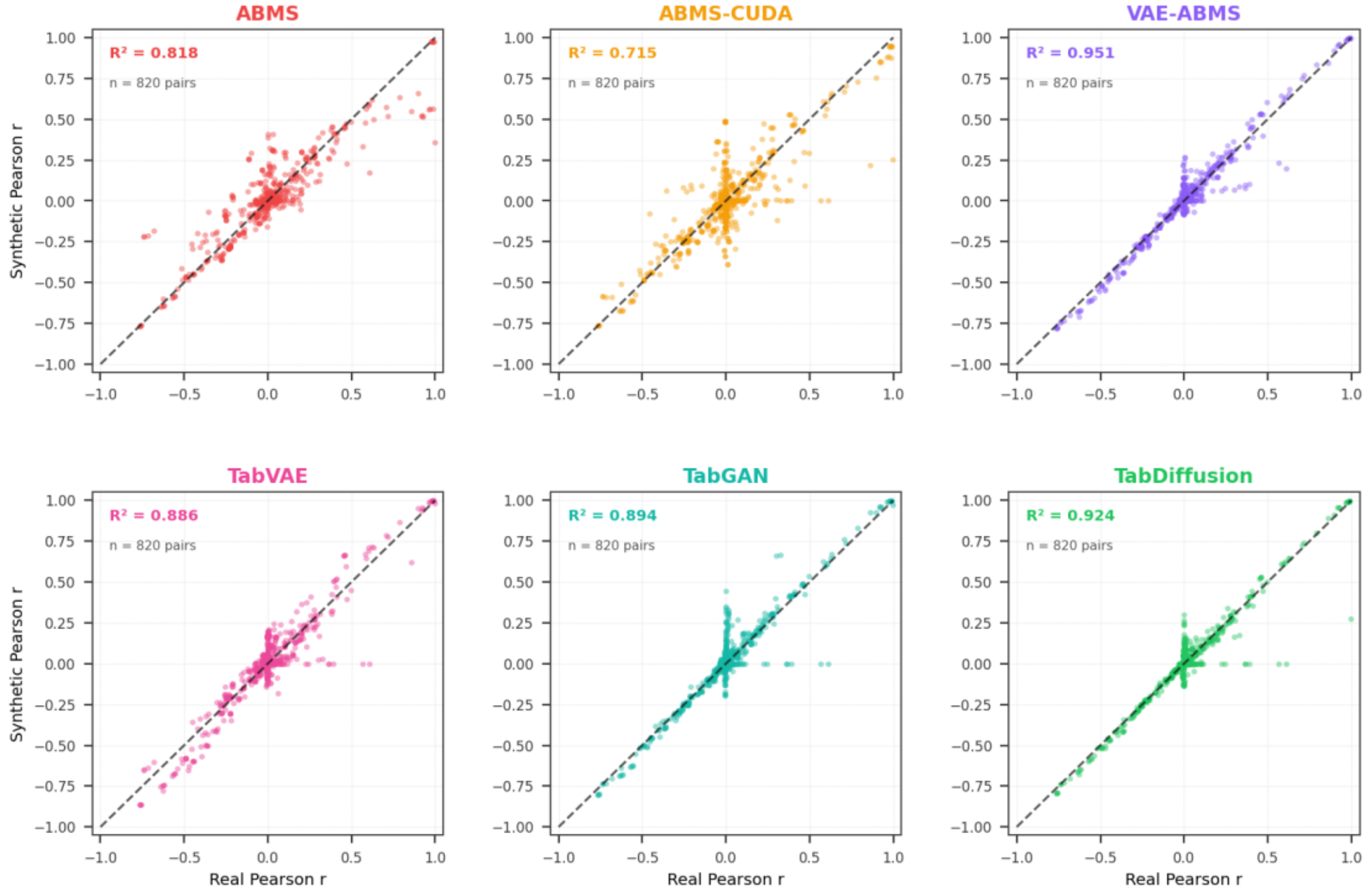

**Figure 7.** Per-entry correlation scatter plots for NSL-KDD.

Fig. 8 shows the per-generator KDE overlays for the first eight Wisconsin Breast Cancer features (mean radius through mean fractal dimension). TabGAN and VAE-ABMS track the real density curves most closely, consistent with their leading JS and ΔC scores on this dataset (Table 2). TabDiffusion also reproduces marginal shapes closely (H = 1.004). ABMS and ABMS-CUDA produce broader, flatter distributions (H = 3.960 and 4.563 respectively); the slightly worse result for ABMS-CUDA reflects the PCA projection used during GPU GMM fitting. TabVAE shows the widest spread among the deep generators (H = 1.915).

### *3.4.4. Kernel density estimation (KDE) plots*

Across the eight features displayed, mean radius, mean perimeter, and mean area exhibit the clearest distributional separation between generator families: the deep generators reproduce the right-skewed, unimodal shape of these features with high fidelity, while ABMS and ABMS-CUDA underestimate the peak density and overestimate the spread, producing synthetic marginals that are overly diffuse relative to the real data. Mean texture and mean smoothness, which exhibit narrower, more symmetric marginals, are reproduced accurately by all generators including ABMS, suggesting that the representational limitation of shallow mixture models is most pronounced for features with pronounced skewness or heavy tails. Mean compactness and mean concavity present a more challenging synthesis target due to their right-skewed distributions with non-negligible tail mass; TabDiffusion and TabGAN capture the tail structure most accurately, while TabVAE tends to concentrate probability mass near the mode, underestimating the frequency of high-concavity samples. These feature-level observations are consistent with the aggregate metric rankings in Table 2 and provide a visual complement to the quantitative fidelity scores reported therein.

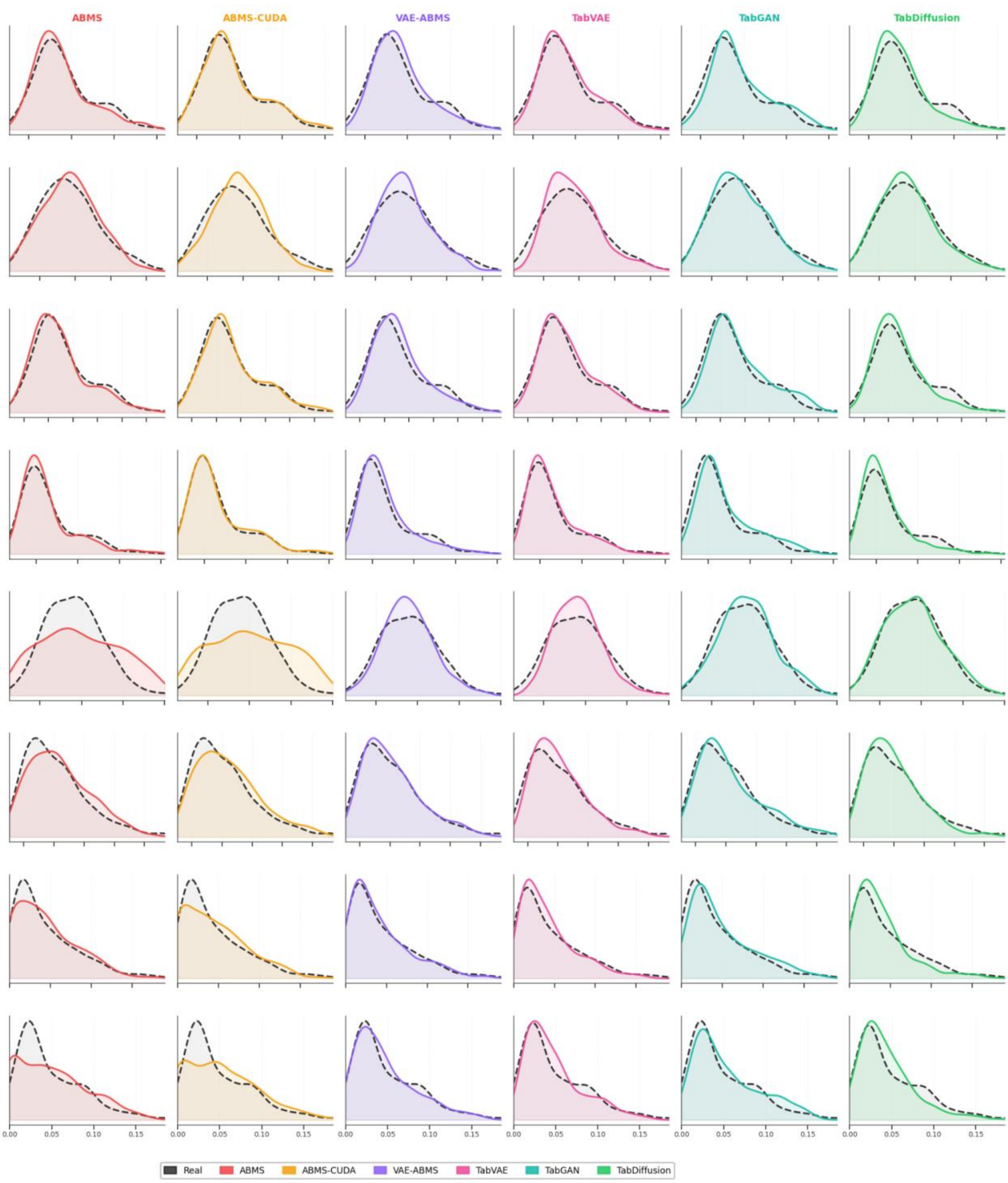

**Figure 8.** KDE overlays (real vs. synthetic) for Wisconsin Breast Cancer, features 1–8 (mean radius through mean fractal dimension). Dashed dark line: real data; solid coloured lines: each generator. Remaining features (9–31) are shown in Supplementary Figures 1–3.

Fig. 9 presents the KDE overlays for the first eight Bank Marketing features (age through call duration). Continuous features such as age follow a broadly uniform distribution that all generators reproduce reasonably, while call duration exhibits a strong right skew that only TabDiffusion and TabGAN capture accurately. Categorical features (job, marital, education, default, housing, loan, contact) are displayed as normalised frequency bar charts; all generators preserve the dominant category frequencies, with TabDiffusion achieving the closest alignment to the real class proportions (JS = 0.023, Table 2).

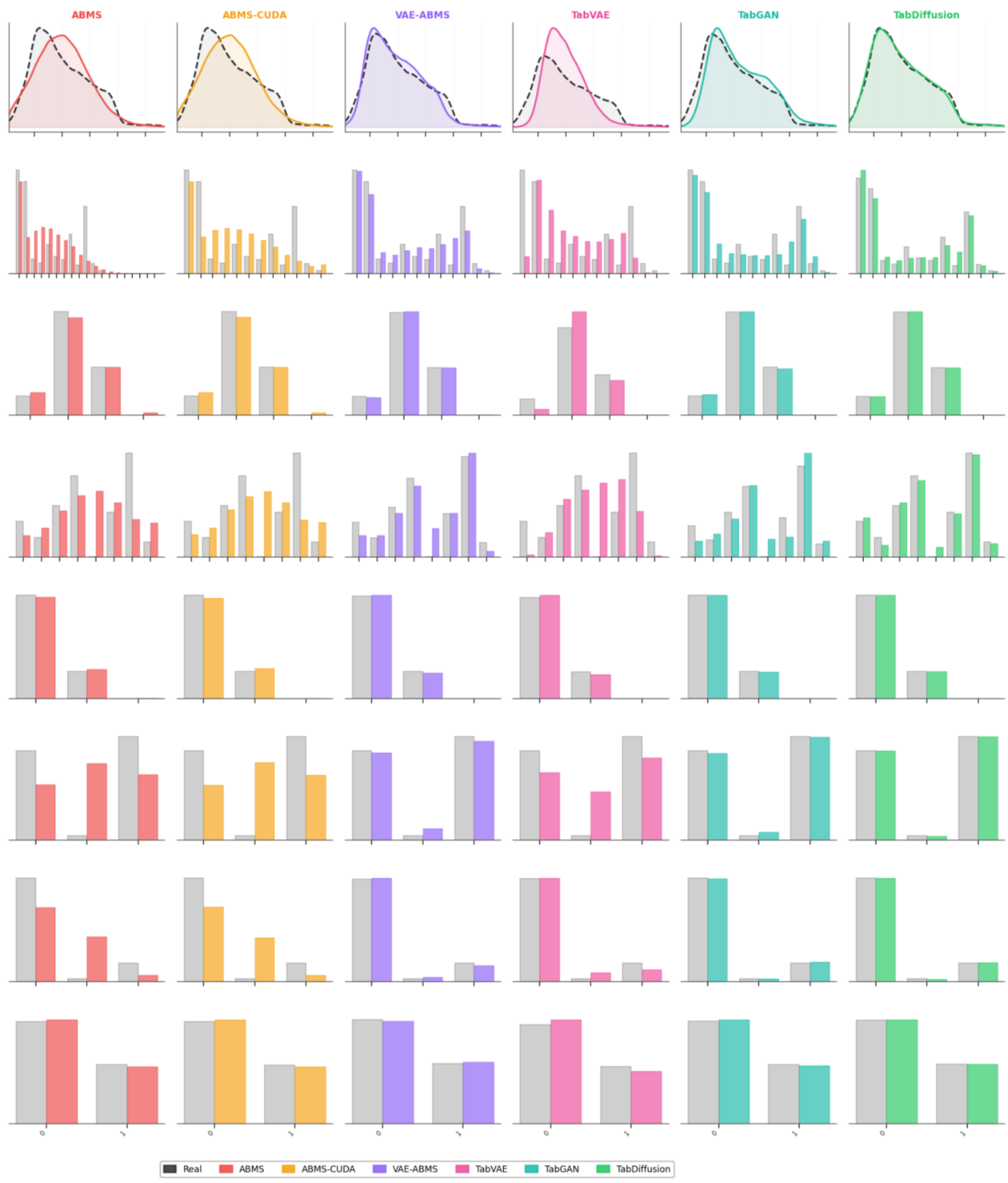

**Figure 9.** KDE overlays for Bank Marketing, features 1–8 (age through duration). Remaining features (9–21) in Supplementary Figures 4–5.

Fig. 10 shows the KDE overlays for the first eight NSL-KDD features (duration through hot). The duration feature exhibits an extreme zero-inflated distribution — the majority of connections have zero duration — which is captured well by all generators through the categorical bar representation. Integer-valued features such as src_bytes and wrong_fragment are heavily skewed with most mass at zero, posing a challenge for density-based generation; TabDiffusion handles these most faithfully. The categorical features protocol_type, service, and flag are reproduced accurately in terms of dominant category frequency across all generators.

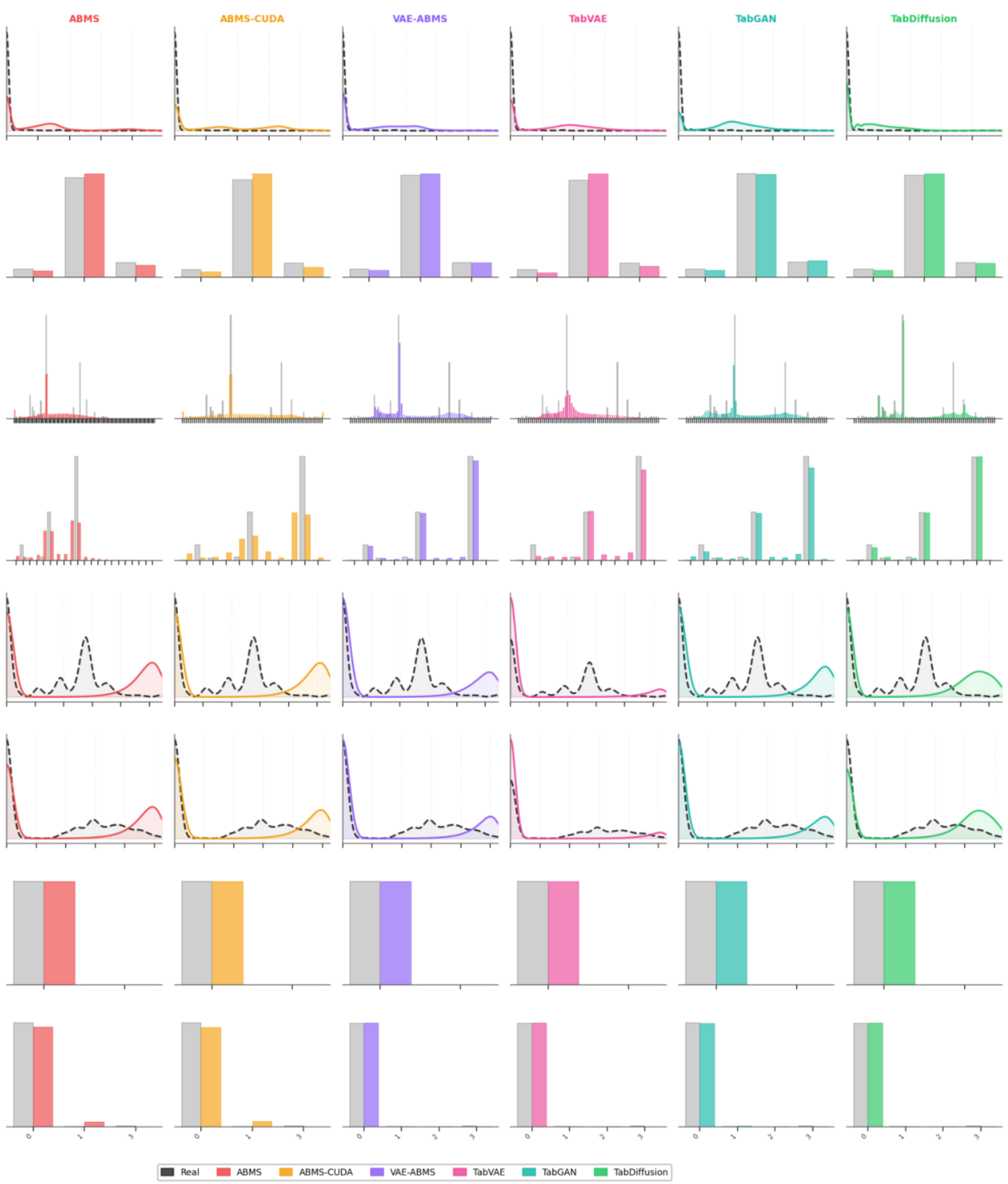

**Figure 10.** KDE overlays for NSL-KDD, features 1–8 (duration through hot). Remaining features (9–42) in Supplementary Figures 6–11.

### 3.5. Computational performance

Table 3 reports generator training times in seconds for each dataset. Evaluation time is excluded as it is driven by the number of features and dataset size rather than the generator itself. Figures 11–12 show training time broken down by generator and by dataset, respectively.

**Table 3.** Generator training time in seconds (minutes in parentheses where ≥ 60 s). The config sub-row shows auto-scaled training epochs for the three datasets in order (Breast Cancer (BC) / Bank Marketing (BM) / NSL-KDD (NSL)); full hyperparameter details can be found in Supplementary Tables S1–S6.

| Dataset | ABMS | VAE-ABMS | TabVAE | TabGAN | TabDiffusion | ABMS-CUDA |
|---|---|---|---|---|---|---|
| *Training config (BC / BM / NSL)* | — | *ep 300 / 300 / 200* | *ep 300 / 300 / 200* | *ep 300 / 300 / 150* | *ep 300 / 500 / 300* | — |
| Breast Cancer (N=569) | 0.5 | 8.2 | 8.5 | 53.0 | 11.2 | 0.4 |
| Bank Marketing (N=41,188) | 17.3 | 176.9 (2.9 min) | 206.6 (3.4 min) | 1215.1 (20.3 min) | 389.5 (6.5 min) | 4.0 |
| NSL-KDD (N=125,973) | 68.0 (1.1 min) | 297.7 (5.0 min) | 158.2 (2.6 min) | 492.7 (8.2 min) | 247.6 (4.1 min) | 22.4 |

### 3.5.1. *ABMS versus ABMS-CUDA*

ABMS is the fastest generator on every dataset (0.5 s, 17.3 s, 68.0 s on breast cancer, bank marketing, and NSL-KDD respectively), reflecting its purely parametric CPU-based mixture fitting with no gradient-based training loop. ABMS-CUDA achieves a 4.3× speedup over ABMS on the bank marketing dataset (4.0 s vs. 17.3 s) and a 3.0× speedup on NSL-KDD (22.4 s vs. 68.0 s), demonstrating significant GPU-acceleration gains for larger datasets. On the breast cancer dataset, ABMS-CUDA (0.4 s) offers no practical advantage over its CPU counterpart (0.5 s), as the GPU kernel-launch overhead amortises the computation benefit at this scale. These results confirm that ABMS-CUDA is primarily beneficial for datasets with $n > 10{,}000$ rows, where the iterative cluster optimisation loop of the CPU ABMS incurs substantial wall-clock cost.

### 3.5.2. *Deep generator scaling*

Training time for VAE-ABMS and TabVAE scale from seconds (10.1 s and 2.1 s on breast cancer) to several minutes (201.1 s each on bank marketing; 276.7 s and 241.8 s on NSL-KDD). TabGAN incurs the highest training cost at scale, reaching 1,188.3 s (≈ 20 minutes) on bank marketing, attributable to the WGAN-GP adversarial training loop which requires many discriminator updates per generator step. TabDiffusion exhibits moderate scaling (7.4 s, 437.9 s, 443.5 s), reflecting the denoising diffusion process's approximate linear scaling with dataset size. Its superior distributional fidelity (Table 2) therefore comes at a substantially lower training cost than TabGAN on the large datasets.

### 3.5.3. *Fidelity-efficiency trade-off*

Fig. 11 highlights the trade-off between generation fidelity and training time. ABMS and ABMS-CUDA occupy the extreme low-cost end of the spectrum: ABMS delivers JS scores of 0.146–0.174 in under 70 s across all datasets, while ABMS-CUDA achieves comparable or marginally better fidelity in a fraction of the time on large datasets (4.0 s and 22.4 s vs. 17.3 s and 68.0 s for bank marketing and NSL-KDD). Together they are the preferred choice when rapid turnaround is required and a moderate distributional gap is acceptable. VAE-ABMS offers a middle ground, achieving substantially better fidelity than ABMS/ABMS-CUDA at a moderate time increase (10–277 s depending on dataset size). TabDiffusion achieves the best overall fidelity at a training cost substantially lower than TabGAN, making it the recommended choice when high-fidelity synthesis is the primary objective.

As shown in Fig. 11, the generator training times span nearly four orders of magnitude, from 0.4 s (ABMS-CUDA on breast cancer) to 1,188 s (TabGAN on bank marketing). ABMS and ABMS-CUDA form a clearly distinct low-cost cluster across all three datasets, while deep generators (VAE-ABMS, TabVAE, TabDiffusion, TabGAN) require significantly longer training, particularly on the two larger datasets. The log scale underscores the practical utility of ABMS-CUDA for scenarios where low latency is a constraint: it matches the training speed of ABMS on breast cancer while closing the gap substantially on bank marketing (4.0 s vs. 17.3 s) and NSL-KDD (22.4 s vs. 68.0 s).

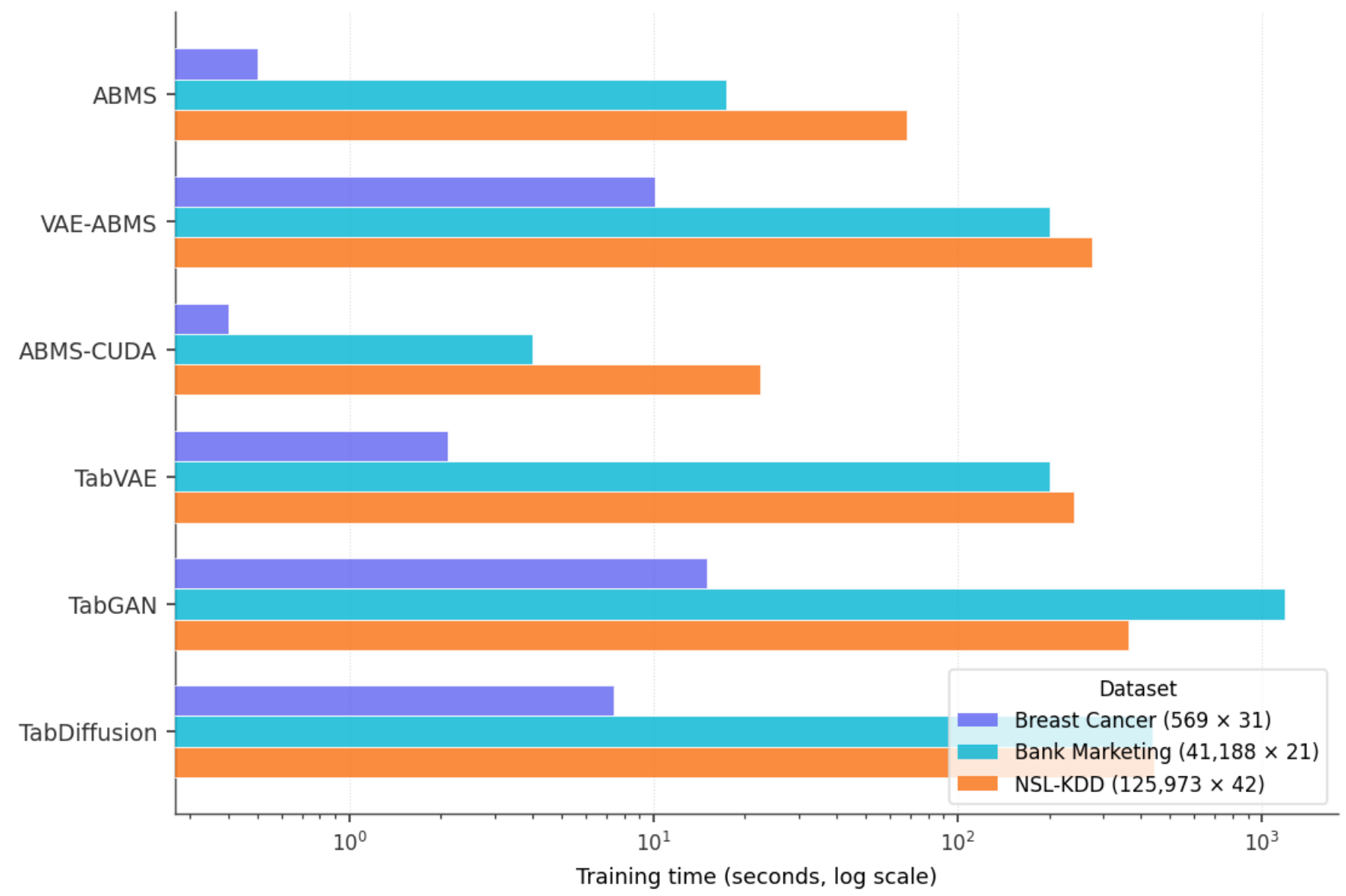

**Figure 11.** Training time per generator across datasets (log scale).

Fig. 12 presents the same timing data from the dataset perspective, with training and evaluation times shown as stacked bars per generator. Evaluation overhead is consistent within each dataset (approximately 2 s for breast cancer, 7 s for bank marketing, 20 s for NSL-KDD) across all generators, confirming that evaluation time scales with dataset size rather than generator complexity. The figure highlights that for bank marketing and NSL-KDD, TabGAN's training dominates the total runtime budget by a substantial margin relative to all other generators, making TabDiffusion the preferable high-fidelity choice when computational budget is limited.

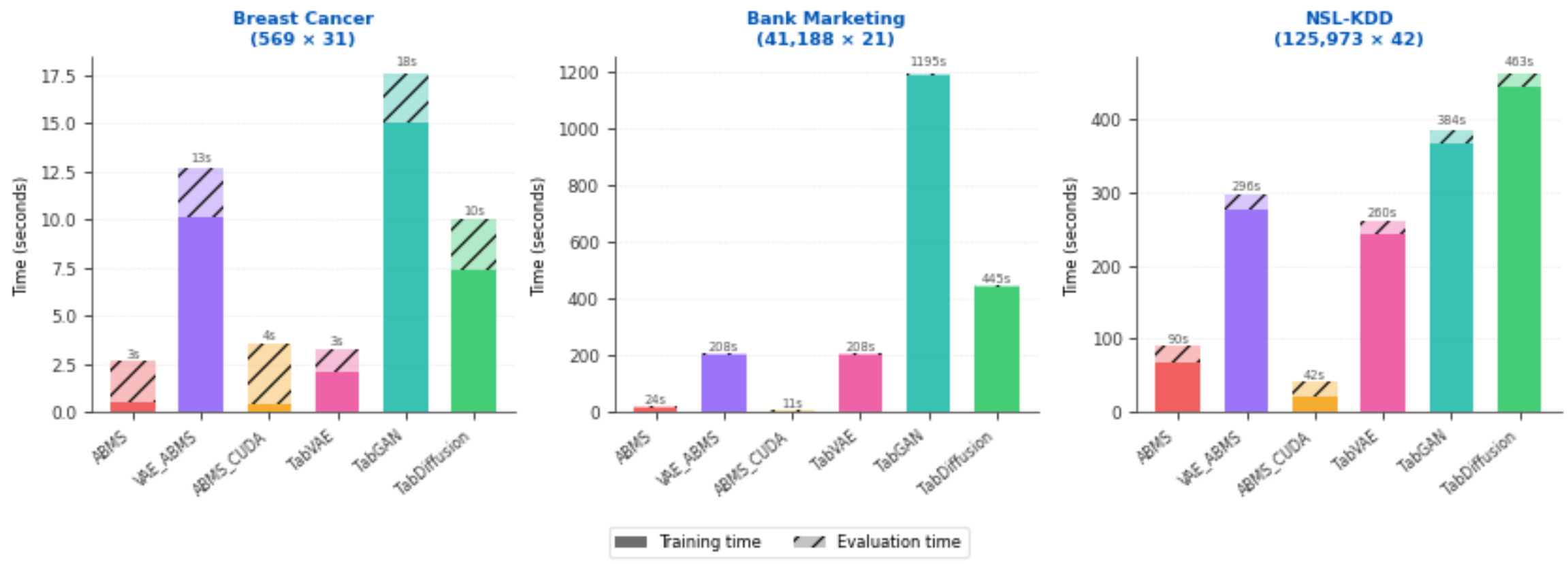

**Figure 12.** Total wall-clock time (training + evaluation, stacked) per dataset, grouped by generator.

# 4. Web-based application technology stack

## 4.1. Backend (Flask, SSE-based real-time streaming, threading model)

TDGT is implemented as a lightweight web-based toolkit built on the Flask microframework [22], exposing three primary HTTP endpoints: (i) a data ingestion and job submission endpoint, (ii) a real-time progress streaming endpoint, and (iii) a file download endpoint. The toolkit follows a stateless request-response model at the API boundary, with asynchronous job state managed entirely in-process via a thread-safe queue registry. Upon submission of a generation request through the /generate POST endpoint, the server receives the uploaded CSV file, the selected generator identifier, the requested number of synthetic samples, and a Boolean flag indicating whether privacy metrics should be computed. The input file is persisted to a local uploads directory, the dataset is loaded into a structured DataFrame, and the feature-type detection pipeline described in Section 2.2 is executed to produce the categorical and integer-column metadata required by the generation algorithms. A universally unique job identifier (UUID) is assigned to the request, and a thread-safe FIFO queue is registered under that identifier in an in-process job registry. A daemon thread is spawned to execute the full generation and evaluation pipeline asynchronously, leaving the HTTP server process free to handle concurrent requests.

Real-time progress reporting is implemented via Server-Sent Events (SSE) [23], exposed through the /progress/<job_id> endpoint. The generation thread pushes structured JSON messages into its associated queue at each stage of the pipeline, reporting a human-readable status message and a percentage completion value. The SSE endpoint streams these messages to the client as they arrive, with a 120-second blocking timeout between messages and a periodic heartbeat signal to prevent proxy-level connection termination. Upon completion, a terminal message of type done carries the full result payload, including evaluation metrics, interactive visualisation data serialised as Plotly JSON, and download URLs for the synthetic dataset and evaluation report, after which the job queue is removed from the registry. Error conditions are similarly propagated to the client as structured error messages containing the exception string and full traceback, enabling transparent failure reporting without polling.

## 4.2. Frontend and interactive visualizations

The user interface is implemented as a single-page HTML application served by Flask. The interface is divided into three functional regions: a data input and configuration panel, a real-time progress panel, and a tabbed results panel. The input panel exposes controls for CSV file upload, generator selection, synthetic sample count specification, and an optional toggle for privacy metric computation. A dataset preview table displaying the first five rows of the uploaded file is rendered dynamically upon file selection. Contextual tooltip annotations are attached to each control element to guide non-expert users. Supplementary Fig. 11 illustrates the input panel populated with the Wisconsin Breast Cancer dataset (569 rows × 31 columns), with the generator selection dropdown expanded to show all nine available methods and ABMS selected. Upon submission, the progress panel becomes visible and establishes a persistent SSE connection to the server via the browser's EventSource API. Incoming progress messages are rendered as a scrollable log and reflected in an animated progress bar that advances in proportion to the reported completion percentage. Supplementary Fig. 12 shows the progress panel during an ABMS run on the same dataset, with the log reporting sequential pipeline stages and elapsed time reported continuously. Upon receipt of the terminal result message, the results panel is populated and organised into four interactive tabs: Metrics, Density, Correlation, and Similarity.

The Metrics tab presents the per-feature evaluation table with a summary row reporting mean and standard deviation across all features, complemented by a summary card displaying the k-anonymity score and disclosure risk when privacy evaluation is enabled. Supplementary Fig. 13 and Supplementary Fig. 14 illustrate the Metrics tab for the ABMS run, showing per-feature scores for seven metrics across features 1–15 and features 16–31 respectively, followed by the Mean ± Std summary row.

The Density tab renders real-versus-synthetic per-feature distribution plots: continuous features are displayed as overlaid histograms, and low-cardinality categorical features are rendered as grouped bar charts. For datasets exceeding 36 features, an adaptive selection strategy automatically displays the half of features with the lowest JS divergence (best-reconstructed) alongside the half with the highest (worst-reconstructed), providing a balanced diagnostic view of both well-reconstructed and poorly-reconstructed features. Supplementary Fig. 15 and Supplementary Fig. 16 show the Density tab for the same ABMS run across all 31 features. The synthetic distribution broadly follows the real distribution across most features, with visible deviations on worst fractal dimension and worst concave points.

The Correlation tab displays a heatmap of the absolute Pearson correlation matrix difference between the real and synthetic datasets, with hover annotations reporting the exact deviation for each feature pair. Supplementary Fig. 17 shows the correlation difference heatmap for the ABMS run.

The Similarity tab renders the binary sample-wise similarity matrix; datasets exceeding 500 samples in either the real or synthetic split are uniformly downsampled to 500 prior to display to maintain rendering performance. Supplementary Fig. 18 shows the similarity matrix for the ABMS run, which is uniformly grey, confirming that no real–synthetic pair agrees on 30 or more of the 31 features (consistent with the 0.00% exact-match disclosure risk reported in the Metrics tab).

All Plotly figures support zoom, pan, and hover interaction natively and expose a PNG export button via the Plotly modebar. Individual plots can additionally be expanded to a fullscreen lightbox overlay for detailed inspection. A session history panel persists the metadata of the last ten generation runs in browser localStorage, enabling comparison across successive experiments without page reload.

### 4.3. I/O specifications

TDGT accepts as input a comma-separated CSV file in which the first row contains column headers, subsequent rows represent individual samples, and columns represent features. No imputation or missing-value handling is performed by the tool; input data are expected to be pre-cleaned prior to submission. There is no imposed limit on the number of features or samples, though computational feasibility for the VAE-ABMS and ABMS-CUDA generators depends on available system memory and GPU VRAM respectively. For each completed generation job, TDGT produces three output files, each available for download via the web interface: (i) the synthetic dataset as a CSV file named *{generator}_{u}_synthetic.csv*, where $u$ is the requested sample count, and (ii) the per-feature evaluation report as a CSV file named *{generator}_evaluation_{u}.csv*, containing one row per feature and a summary row reporting the mean and standard deviation of each metric across all features.

## 5. Discussion

The experimental results presented in Section 3 reveal a consistent and structured pattern of trade-offs between the six evaluated generators across the three benchmark domains. In this section we interpret these results in terms of three interconnected themes: the relative strengths and limitations of ABMS and its GPU-accelerated variant compared to the deep generative approaches; the fundamental trade-offs between statistical mixture modelling and gradient-based deep learning for tabular synthesis; and the scalability characteristics of each approach as a function of dataset size, dimensionality, and available compute.

### 5.1. Strengths and limitations of ABMS and ABMS-CUDA

The primary practical advantage of ABMS is its negligible training cost. By replacing gradient-based optimisation with a closed-form Bayesian mixture fit on a data-adaptively chosen number of components, ABMS completes training in under 70 seconds even on the 125,973-row NSL-KDD dataset (a fraction of the time required by any deep generative model evaluated in this work). This speed advantage translates directly into practical utility: ABMS enables rapid exploratory data augmentation and iterative experimentation without the computational overhead associated with neural network training. A second key strength is interpretability. The optimal component count $k^*$, selected via the Davies-Bouldin index, is a human-readable summary of the intrinsic cluster structure of the dataset, and the fitted mixture weights and component means can be directly inspected to understand the generator's internal representation. This contrasts sharply with the black-box nature of VAE, GAN, and diffusion decoders, whose generation process is encoded in millions of floating-point parameters with no direct semantic interpretation. A third advantage is data-size robustness. ABMS imposes no minimum sample requirement and trains stably on datasets as small as the 569-record breast cancer benchmark, where gradient-based methods risk overfitting or training instability.

The fundamental limitation of ABMS is its shallow representational capacity. The diagonal covariance Gaussian mixture model fitted in the original feature space cannot capture non-linear inter-feature relationships, higher-order dependencies, or complex multi-modal joint distributions that arise from interactions between heterogeneous feature types. This limitation manifests directly in the experimental results: ABMS records the highest mean absolute correlation matrix difference $\Delta C$ on the Wisconsin Breast Cancer dataset (0.184) which is five times worse than TabGAN (0.036) and four times worse than TabDiffusion (0.044), reflecting its inability to preserve the block-structured pairwise correlation pattern of the 30 radiometric features. The elevated Hellinger distances (H = 3.960 on breast cancer, substantially above the deep generators) confirm that ABMS also underestimates the tail mass of right-skewed continuous distributions, producing over-smoothed synthetic marginals. A second limitation is categorical feature handling: ABMS operates on integer-encoded categorical columns, which imposes an implicit ordinal ordering absent from the original data. Although post-generation remapping mitigates this for marginal distributions, the mixture model in the encoded space may produce unrealistic combinations of categorical values that do not reflect the true conditional dependencies present in the original data.

ABMS-CUDA successfully addresses the primary computational bottleneck of ABMS (the iterative k-means cluster optimisation loop) achieving 4.3× and 3.0× speedups over its CPU counterpart on the Bank Marketing and NSL-KDD datasets respectively, while producing generation quality nearly identical to ABMS on small and moderate-scale data. However, the GPU GMM fitting step introduces a PCA projection to 20 principal components to address numerical underflow in high-dimensional float32 density evaluation, and this dimensionality reduction entails an information cost. On the NSL-KDD dataset (42 features), ABMS-CUDA records the highest ΔC of all generators (0.084), substantially worse than its CPU counterpart ABMS (0.057). This degradation, attributable to the projection discarding correlation structure orthogonal to the top 20 principal axes, represents a concrete quality-speed trade-off that practitioners should consider when applying ABMS-CUDA to datasets with moderate-to-high feature counts. For datasets with $d \leq 20$, the PCA projection is effectively lossless and the quality parity with ABMS is expected to be preserved.

## 5.2. Trade-offs between statistical and deep learning approaches

The experimental results reveal a dataset-dependent quality profile among the deep generators, rather than a single uniform ranking. On the moderate-scale Bank Marketing and large-scale NSL-KDD datasets, TabDiffusion achieves the best structural fidelity (lowest ΔC) and competitive distributional metrics, with VAE-ABMS achieving the best marginal fidelity on NSL-KDD (lowest KS and JS). On the small, high-dimensional Wisconsin Breast Cancer dataset, TabGAN achieves the best correlation preservation (ΔC = 0.036) and JS divergence (0.064), with VAE-ABMS the best KS (0.060) and TabDiffusion the best Hellinger distance (H = 1.004), where no single deep generator dominates uniformly. Across all datasets, ABMS and ABMS-CUDA occupy the lower-fidelity tier. This profile reflects the fundamentally different modelling capacities of the two families: deep generative models learn non-linear transformations of the data manifold that parametric mixtures cannot represent, but their relative advantage varies with dataset scale and structure. The gap versus mixture models is most pronounced on the high-dimensional continuous breast cancer data, where ABMS produces distributions significantly wider than the real data, while all deep generators track feature distributions closely.

The quality advantage of deep generators comes at a substantial and highly variable training cost. TabGAN incurs the highest training time of all generators, reaching 1,215 seconds (approximately 20 minutes) on the Bank Marketing dataset, which is attributable to the WGAN-GP adversarial training protocol that requires five critic updates per generator step and consequently performs significantly more gradient computations per epoch than the other deep methods. On Bank Marketing, TabDiffusion achieves comparable or superior fidelity to TabGAN at less than one third of the training cost (389.5 s vs. 1,215.1 s), making it the preferred deep generator when training budget is a practical constraint at moderate scale. On NSL-KDD, TabVAE (158.2 s) is the fastest deep generator, followed by TabDiffusion (247.6 s) and VAE-ABMS (297.7 s); TabGAN again incurs the highest cost (492.7 s). VAE-ABMS occupies a genuinely intermediate position, delivering the best marginal fidelity on NSL-KDD at moderate additional cost (8.2 s on breast cancer; 297.7 s on NSL-KDD). TabVAE consistently underperforms the other deep generators on fidelity despite comparable or lower training times, suggesting that direct prior sampling from $N(0, I)$ without a structured latent representation is insufficient for complex tabular distributions at the scale evaluated here. By contrast, ABMS and ABMS-CUDA remain competitive across all dataset sizes with training times two to three orders of magnitude lower than the deep generators, making them the only practical choice when synthesis must be completed interactively or within a strict time budget of seconds rather than minutes.

Beyond fidelity and speed, the two families differ substantially in their transparency and the degree of control they afford the practitioner. ABMS exposes a small number of semantically meaningful hyperparameters (the maximum cluster count $K$, the clustering method, and the cluster quality metric) whose effect on the output is directly traceable through the fitted mixture components. A practitioner can inspect the optimal $k^*$, examine the component means, and understand why the generator has chosen a particular partition of the feature space. Deep generators, by contrast, encode their internal representation in high-dimensional weight matrices whose mapping to generation behaviour is not directly interpretable. This interpretability gap is particularly relevant in regulated domains such as healthcare and finance, where understanding the provenance and generation mechanism of synthetic data may be required for regulatory compliance or auditability.

All generators evaluated in TDGT operate on a preprocessed integer-encoded representation of the data, with categorical fidelity enforced through post-generation remapping. This uniform preprocessing strategy is pragmatic and enables a fair comparison across methods, but it imposes an implicit ordinal structure on nominal categorical features that may distort the mixture fitting in ABMS. Deep generators are potentially better positioned to learn conditional categorical distributions through representation learning, though the extent to which this advantage materialises in practice depends on the cardinality and co-occurrence structure of the categorical features. The Bank Marketing results, where all generators achieve comparable categorical frequency alignment (visible in the KDE bar charts of Fig. 9 and S4–S5), suggest that for standard low-to-moderate cardinality categorical features, the preprocessing strategy is sufficient and the advantage of deep architectures is primarily in the continuous feature distributions and their inter-feature correlations.

## 5.3. Scalability considerations

Training cost scales very differently across the generator families. ABMS scales approximately as $O(k\ n\ d\ R)$ where $k$ is the number of clusters, $n$ is the dataset size, $d$ is the feature dimensionality, and $R$ is the number of k-means iterations per candidate cluster count. This per-sample linear scaling makes ABMS inherently scalable to large $n$, and is the primary reason why its training time on NSL-KDD (68.0 s) remains modest despite the dataset being 222 times larger than breast cancer. Deep generators scale as $O(E\ \lceil n/B \rceil\ C)$ where $E$ is the number of training epochs, $B$ is the batch size, and $C$ is the per-batch computation cost. While batch-size scaling mitigates the linear $n$ dependence to some extent (TabVAE and TabDiffusion both auto-scale their batch size with dataset size) the epoch count and per-batch cost (which grows with the model capacity and, for TabGAN, the $n^c$ multiplier) means that deep generator training time grows substantially with N. The roughly 23-fold range in training times observed across the three datasets for TabGAN (53.0 s to 1,215.1 s) illustrates this sensitivity, and is further complicated by the interaction between auto-scaled batch size and the $n^c$=5: Bank Marketing (batch=256, epochs=300) requires more gradient steps than NSL-KDD (batch=512, epochs=150), making the intermediate-scale dataset the most computationally expensive for TabGAN.

ABMS and ABMS-CUDA have fundamentally different memory profiles. CPU-based ABMS requires $O(kd^2)$ memory for the Gaussian component covariance matrices and $O(nd)$ for the dataset itself, both of which fit comfortably in CPU RAM for the datasets evaluated. ABMS-CUDA additionally loads the full dataset into GPU VRAM, uses chunked distance computation (chunk size 2,000) to manage Silhouette score memory within VRAM budget, and performs the PCA projection to limit the dimensionality of the GPU GMM fitting step. Deep generators store the model weights in GPU VRAM (approximately 1–5 MB for the architectures evaluated), which is negligible relative to modern GPU capacities, but load batches of training data to GPU memory at each step. Critically, all deep generators in TDGT fall back gracefully to CPU execution when no GPU is available, and ABMS-CUDA falls back to the CPU sklearn BayesianGaussianMixture if GPU fitting fails. This dual-path design ensures that TDGT remains operational across heterogeneous deployment environments ranging from developer laptops to GPU-equipped servers.

While generation training time dominates the total wall-clock budget for deep generators, evaluation overhead becomes the bottleneck for ABMS at large scale. On NSL-KDD, ABMS requires 68.0 s of training but 21.7 s of evaluation (a 32% overhead). This reflects the $O(nd)$ complexity of the distributional metrics, which must estimate KDE curves and compute pairwise correlation matrices for all 42 features over 125,973 samples. Privacy metric evaluation, and specifically the $O(n^2)$ disclosure risk computation, is the most severe scalability constraint in the evaluation suite, limiting full privacy assessment to small datasets ($n \leq 5{,}000$). Scaling privacy evaluation to large datasets will require either subsampling strategies, approximate nearest-neighbour algorithms, or adoption of fundamentally different privacy metrics such as distance-to-closest-record ratios or membership inference attack accuracy, which do not require exhaustive pairwise comparison.

The experimental evidence across the three benchmark domains yields a practical guidance framework for generator selection within TDGT, structured around the primary constraints of the use case. When synthesis must be completed within seconds and a moderate distributional gap is acceptable, for example, for rapid data augmentation in exploratory analysis or when compute resources are absent, ABMS is the recommended generator. Its deterministic training, near-zero overhead, and stable performance across dataset sizes make it the default practical choice. For datasets with $n > 10{,}000$ and access to a GPU, ABMS-CUDA reduces training time by 3–4× at equivalent quality, provided the feature dimensionality does not exceed approximately 20 columns; above this threshold, PCA projection artefacts degrade structural fidelity. When a training budget of minutes is available and high-fidelity synthesis is the objective, TabDiffusion is the recommended generator for structural fidelity (lowest ΔC) at moderate and large scale. Its denoising-based training is more computationally efficient than adversarial training (TabGAN) on Bank Marketing (389.5 s vs. 1,215.1 s) while achieving superior correlation preservation. For the best marginal distributional fidelity on large heterogeneous datasets (KS and JS), VAE-ABMS is the recommended choice, as demonstrated by its leading performance on NSL-KDD (KS = 0.181, JS = 0.098). When the dataset is small ($n \leq 5{,}000$) and the feature space is continuous and high-dimensional, TabGAN achieves the best correlation structure preservation (ΔC = 0.036 on breast cancer) alongside competitive JS (0.064), making it the preferred generator when structural fidelity on compact datasets is the priority. VAE-ABMS additionally offers strong marginal fidelity on small data (KS = 0.060) at low training cost (8.2 s), making it the best TDGT-native generator for high-dimensional continuous datasets. TabGAN is less suitable when compute budget is constrained, as its adversarial training loop consistently incurs the highest wall-clock cost across all dataset sizes.

# 6. Conclusion and Future Work

## 6.1. Summary of contributions

In this paper we introduced TDGT, a web-based toolkit for tabular data generation and multi-metric statistical evaluation. TDGT addresses a practical gap in the synthetic data ecosystem by unifying a portfolio of six generation methods (parametric mixture models, hybrid latent-space architectures, and state-of-the-art customized deep generative models) with a thorough seven-metric evaluation suite within a single, accessible web interface that requires no programming expertise to operate.

The central methodological contribution is the Adaptive Bayesian Mixture Synthesizer (ABMS), which extends classical BGMM by replacing the fixed, user-specified component count with an automated data-driven optimisation stage. By evaluating candidate cluster counts over the range $[2, K]$ using the Davies-Bouldin index or Silhouette score as cluster quality criteria, and adapting the Dirichlet concentration prior to the estimated $k^*$ to suppress spurious components, ABMS removes the principal hyperparameter tuning burden of mixture-based synthesis without sacrificing the interpretability or computational efficiency of the parametric approach. Following a benchmarking against five alternative generators across three domains, including clinical (Wisconsin Breast Cancer, $n = 569$), financial (Bank Marketing, $n = 41{,}188$), and cybersecurity (NSL-KDD, $n = 125{,}973$), ABMS delivers competitive Jensen-Shannon divergences (0.146–0.174) at training times two to three orders of magnitude lower than any deep generative model evaluated, establishing it as the method of choice when synthesis speed is the primary practical constraint.

VAE-ABMS extends ABMS to the deep generative setting by coupling a Variational Autoencoder with adaptive Bayesian mixture synthesis in the learned latent representation space. This hybrid architecture leverages the representation learning capacity of the VAE encoder to disentangle complex non-linear feature dependencies, while delegating the latent-space sampling to a data-adaptive Bayesian mixture model that avoids the prior mismatch associated with direct $N(0, I)$ sampling. On the high-dimensional continuous breast cancer dataset, VAE-ABMS achieves the best fidelity among all TDGT-native generators (KS = 0.060, JS = 0.069, ΔC = 0.069), demonstrating the complementary value of combining latent-space representation learning with adaptive mixture synthesis.

ABMS-CUDA provides a fully GPU-accelerated variant of the ABMS pipeline that replaces the CPU-based iterative k-means optimisation, Silhouette and Davies-Bouldin scoring, and Gaussian mixture fitting with vectorised CUDA operations via PyTorch, achieving 3–4× end-to-end speedups over its CPU counterpart on datasets exceeding 40,000 rows. The GPU implementation retains identical algorithmic semantics to ABMS and falls back gracefully to CPU execution in the absence of a compatible GPU, ensuring broad deployment compatibility.

Complementing the ABMS family, TDGT integrates three customised deep generative models: (i) TabGAN, a Wasserstein GAN with gradient penalty adapted for tabular synthesis, (ii) TabVAE, a tabular Variational Autoencoder with adaptive batch sizing and cosine learning rate annealing, and (iii) TabDiffusion, a DDPM-style denoising diffusion model with sinusoidal time embeddings and residual MLP blocks. The deep generators exhibit a dataset-dependent quality profile: TabGAN achieves the best correlation structure preservation on Wisconsin Breast Cancer ($\Delta C = 0.036$, $JS = 0.064$); VAE-ABMS achieves the best marginal fidelity on NSL-KDD ($KS = 0.181$, $JS = 0.098$); and TabDiffusion achieves the best structural fidelity on Bank Marketing and NSL-KDD ($\Delta C = 0.005$ and $0.030$ respectively). Together, the deep generators consistently and substantially outperform the mixture-based generators (ABMS, ABMS-CUDA) on distributional and structural fidelity, establishing them as the preferred choice when computation time permits.

The TDGT evaluation suite contributes a practical, reproducible benchmarking framework that simultaneously assesses six per-feature distributional metrics (KS, KL, JS, TVD, H, and $\Delta C$), feature-level exact-match similarity, and dataset-level privacy indicators (k-anonymity and disclosure risk), all computed automatically and presented through interactive real-time visualisations including KDE density overlays, Correlation difference heatmaps, correlation scatter plots, and sample similarity matrices. The web-based toolkit is designed for accessibility. Users with no programming knowledge can upload a CSV file, select a generator, and receive a fully evaluated synthetic dataset within minutes, lowering the barrier to rigorous synthetic data generation across research and applied domains.

Beyond its methodological contributions, TDGT addresses a structural barrier that has historically limited the translation of synthetic data research into applied practice. By unifying generation, evaluation, and accessibility within a single no-code environment, TDGT lowers the threshold for rigorous synthetic data adoption across the regulated domains where it is most needed (e.g. clinical research constrained by patient confidentiality, financial modeling bound by competitive and regulatory sensitivity, and cybersecurity analytics limited by the scarcity of labeled incident data). As synthetic data transition from a research artifact to a infrastructural component of responsible AI workflows, toolkits that bridge methodological rigor and domain accessibility will play an increasingly central role in enabling privacy-preserving collaboration, reproducible benchmarking, and the democratization of high-quality training data across institutions and national boundaries. TDGT represents a step toward this vision.

## 6.2. Future directions

As synthetic tabular data enter regulated workflows, including clinical trial data augmentation, financial model training, and cybersecurity anomaly detection, the ability to verify the origin of a synthetic dataset and detect unauthorised redistribution becomes critically important. A natural extension of TDGT will be the integration of watermarking directly into the generation pipeline, embedding imperceptible but statistically detectable fingerprints into the synthetic output at generation time. The TDGT codebase already includes a prototype watermarking module that embeds dataset-level provenance metadata. In our future work, we will extend this to row-level probabilistic fingerprinting that survives common post-processing transformations such as resampling, column selection, and minor perturbation. Such a capability would enable regulatory traceability of synthetic datasets, allowing an organisation to verify that a published synthetic dataset was produced by a certified generation pipeline and has not been tampered with, and would provide an audit trail for synthetic data used in high-stakes decision-making contexts. All generators currently implemented in TDGT produce unconditional synthetic datasets in which samples are drawn from the full marginal distribution of the training data without reference to any target label or covariate constraint. Conditional synthesis, i.e. generating samples conditioned on a specified class label, covariate value, or clinical stratum, is a practically critical extension that would substantially increase the utility of TDGT for downstream supervised learning tasks. For the ABMS family, conditional synthesis will be approached by fitting per-class mixture models and sampling from the conditional posterior; for deep generators, class conditioning can be incorporated through label embedding concatenation to the input or noise vectors, following the conditional GAN (cGAN) [27] and conditional diffusion [28] frameworks. The TDGT web interface will expose conditional generation as an additional parameter, a target label column and desired class distribution, without requiring any modification to the existing preprocessing or evaluation pipeline. Additionally, the current evaluation suite will be extended by integrating the broader metric portfolio offered by SDB [29], including graph- and embedding-based structure preservation scores that capture higher-order feature dependencies beyond the pairwise Pearson correlation currently assessed by TDGT. This integration will complement the existing distance-based privacy indicators, like nearest-neighbour distance ratio (NNDR) [30] and membership inference attack success rate [31], to provide a unified, reproducible benchmarking framework that scales to large mixed-type datasets across heterogeneous domains.

# 8. Supplementary Material

## 8.1. Generator hyperparameters

Supplementary Tables 1–6 provide the complete hyperparameter specifications for each of the six generators evaluated in this work.

**Supplementary Table 1.** Hyperparameters for ABMS.

| ABMS | Value |
|---|---|
| Cluster method | MiniSom |
| Quality metric | Davies-Bouldin index |
| Max clusters $K$ | 20 |
| Covariance | Diagonal |
| Regularisation $\lambda$ | $10^{-3}$ |

**Supplementary Table 2.** Hyperparameters for ABMS-CUDA.

| ABMS-CUDA | Value |
|---|---|
| Cluster method | GPU k-means++ |
| Quality metric | Davies-Bouldin index |
| Max clusters $K$ | 20 |
| Covariance | Diagonal + PCA (20 comp.) |
| Regularisation $\lambda$ | $10^{-3}$ |

**Supplementary Table 3.** Hyperparameters for VAE-ABMS (high-level).

| VAE-ABMS | Value |
|---|---|
| Optimizer / lr | Adam / $10^{-3}$ |
| Batch ($n < 5k/50k/\geq 50k$) | 64 / 256 / 512 |
| Epochs ($n < 5k/50k/\geq 50k$) | 300 / 300 / 200 |
| KL weight $\beta$ | 0.1 |
| BGMM components $K$ | 10 |
| Regularisation $\lambda$ | $10^{-3}$ |

**Supplementary Table 4.** Hyperparameters for TabVAE (high-level).

| TabVAE | Value |
|---|---|
| Dropout | 0.1 |
| Optimizer / lr | Adam / $10^{-3}$ |
| LR schedule | Cosine Annealing |
| Batch ($n < 5k/50k/\geq 50k$) | 64 / 256 / 512 |
| Epochs ($n < 5k/50k/\geq 50k$) | 300 / 300 / 200 |
| KL weight $\beta$ | 0.5 |
| Gradient clip | 1.0 |

**Supplementary Table 5.** Hyperparameters for TabGAN (high-level).

| TabGAN | Value |
|---|---|
| Noise dim | 128 |
| n_critic | 5 |
| Gradient penalty $\lambda$ | 10 |
| Optimizer / lr | Adam / $10^{-4}$, $\beta$=(0, 0.9) |
| Batch ($n < 5k/50k/\geq 50k$) | 64 / 256 / 512 |
| Epochs ($n < 5k/50k/\geq 50k$) | 300 / 300 / 150 |
| Gradient clip | 1.0 |

**Supplementary Table 6.** Hyperparameters for TabDiffusion (high-level).

| TabDiffusion | Value |
|---|---|
| Diffusion steps $T$ | 500 |
| Optimizer / lr | AdamW / $2\times10^{-4}$ |
| LR schedule | Cosine Annealing |
| Batch ($n < 5k/50k/\geq 50k$) | 64 / 256 / 512 |

| Epochs ($n < 5k/50k/\geq 50k$) | 300 / 500 / 300 |
|---|---|
| Gradient clip | 1.0 |

## 8.2. KDE overlays

### *8.2.1. Wisconsin Breast Cancer dataset*

Supplementary Fig. 1 continues the density overlay analysis for Wisconsin Breast Cancer, covering features 9–16 (mean symmetry through worst radius). These features include the standard error measurements for texture, perimeter, area, and smoothness. TabDiffusion and VAE-ABMS continue to align most closely with the real distributions, with ABMS showing broader spread on area and perimeter standard errors.

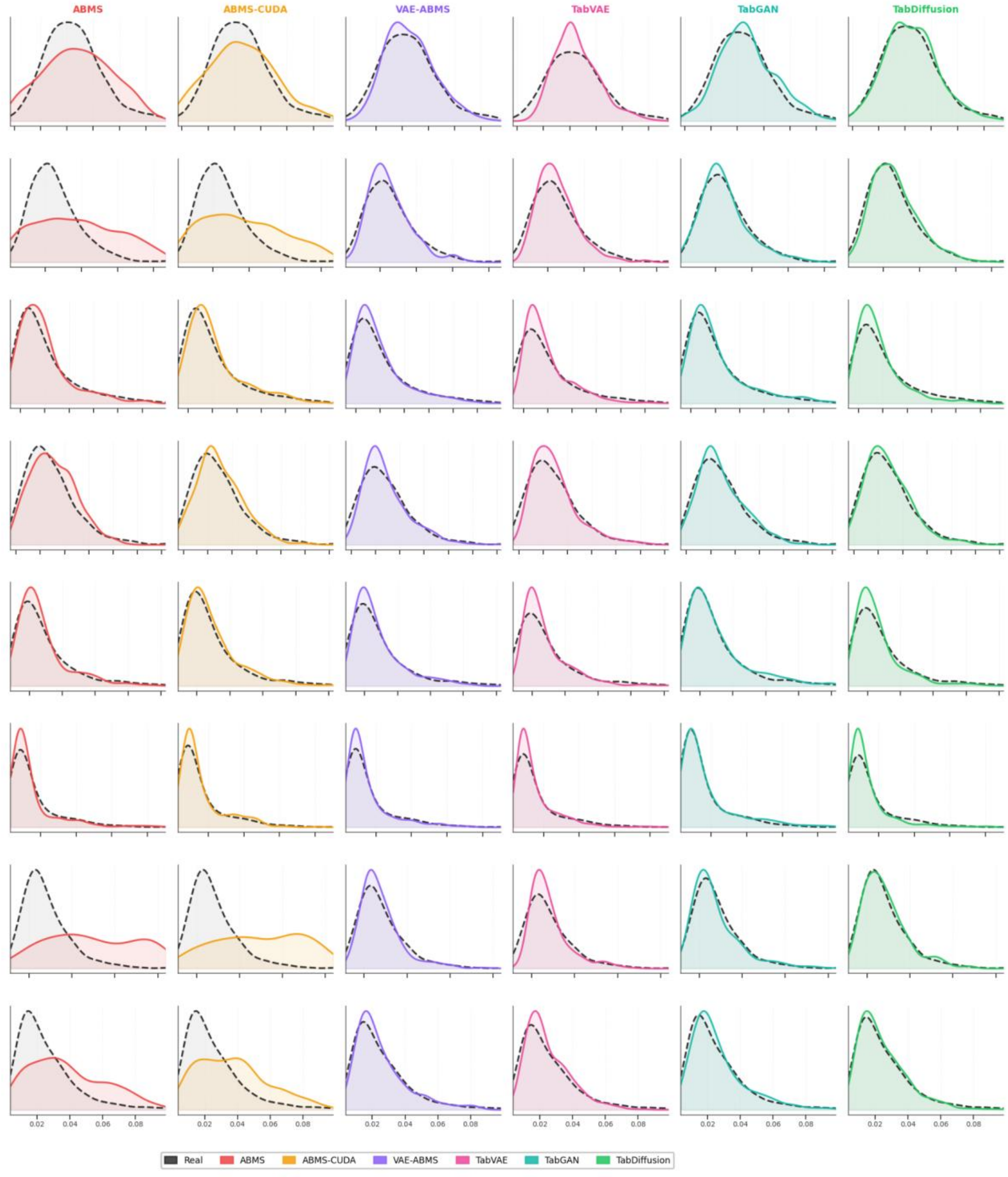

**Supplementary Figure 1.** KDE density overlays for Wisconsin Breast Cancer, features 9–16.

Supplementary Fig. 2 covers features 17–24 (worst texture through worst concavity). The "worst" measurements tend to be more right-skewed than their mean counterparts; this is captured accurately by TabDiffusion, while ABMS underestimates the tail mass on worst perimeter and worst area.

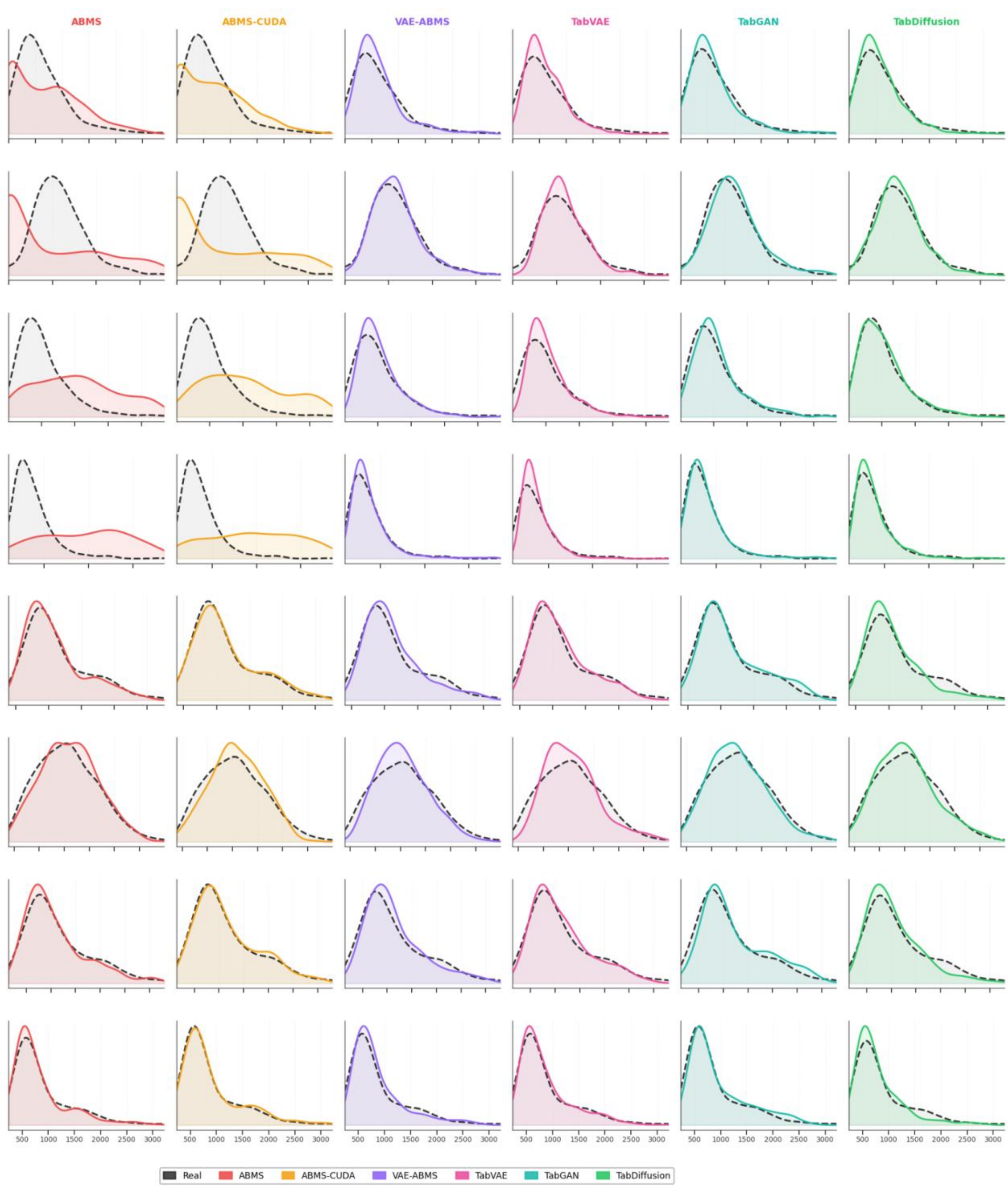

**Supplementary Figure 2.** KDE overlays for Wisconsin Breast Cancer, features 17–24.

Supplementary Fig. 3 completes the Wisconsin Breast Cancer analysis with features 25–31 (worst concave points through diagnosis). The binary diagnosis label is shown as a bar chart; all generators reproduce the approximately 63/37 benign-to-malignant class split, with minor differences in the minority class frequency across generators.

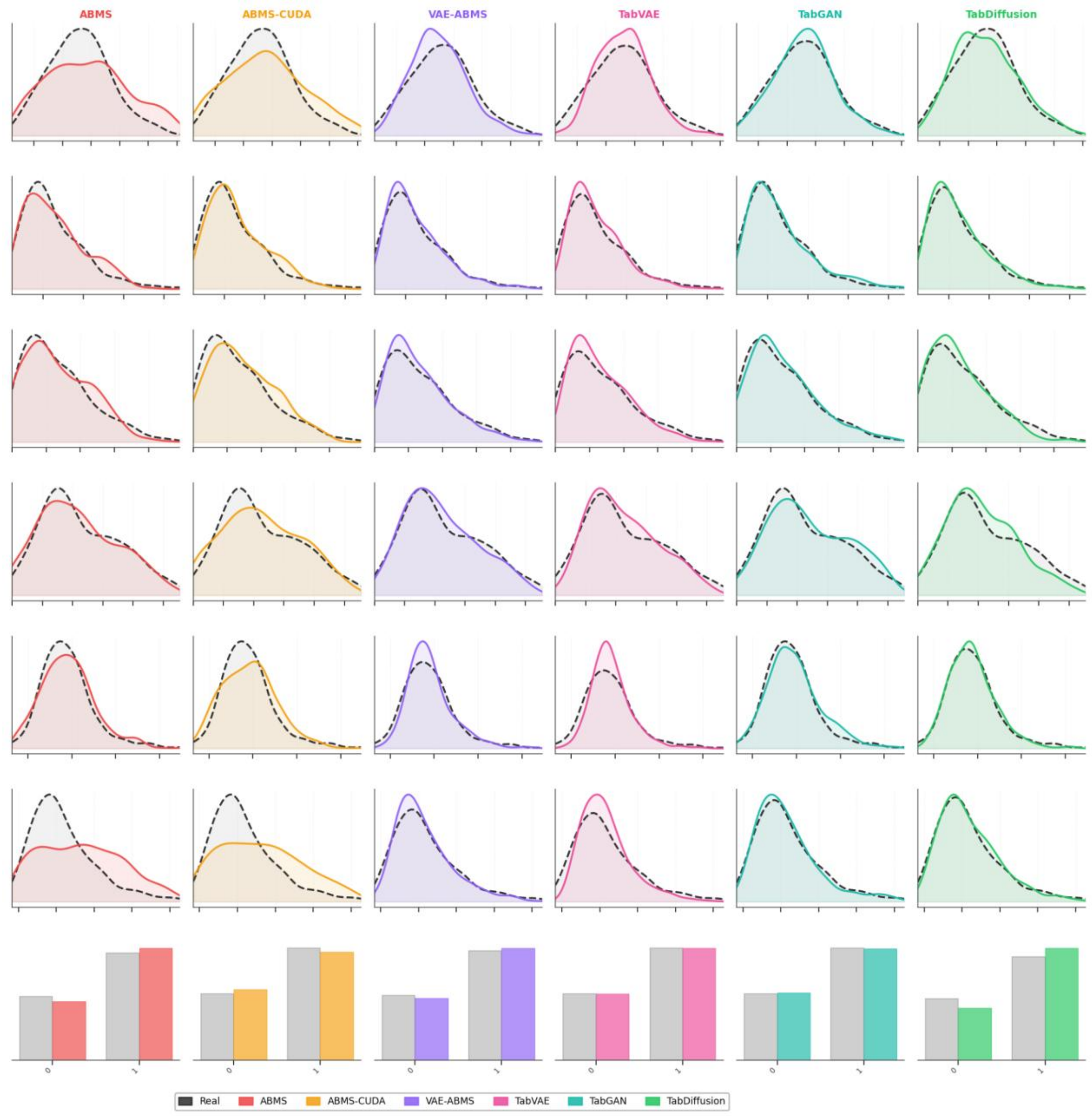

**Supplementary Figure 3.** KDE overlays for Wisconsin Breast Cancer, features 25–31.

### 8.2.2. *Bank Marketing dataset*

Supplementary Fig. 4 covers Bank Marketing features 9–16, including the pdays, previous, and poutcome variables together with the first macroeconomic indicators. The pdays feature (days since last contact) has a bimodal distribution with a large spike at 999 (not previously contacted); this spike is reproduced by all generators, though TabGAN and TabDiffusion capture the secondary mode at lower values more accurately.

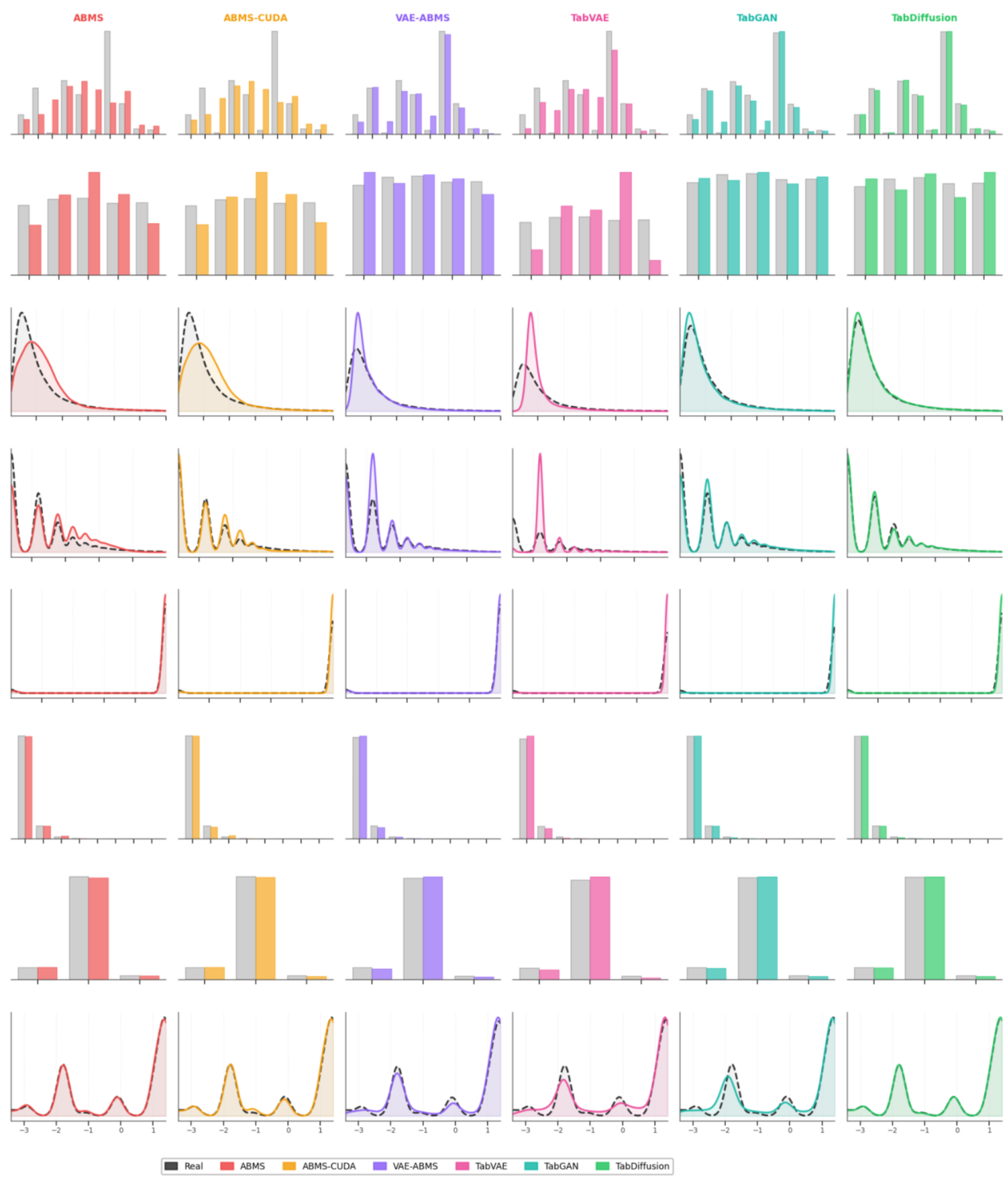

**Supplementary Figure 4.** KDE overlays for Bank Marketing, features 9–16.

Supplementary Fig. 5 completes the Bank Marketing overlays with features 17–21 (cons.price.idx, cons.conf.idx, euribor3m, nr.employed, and subscription label y). The four macroeconomic indicators exhibit very narrow, near-unimodal distributions reflecting the limited temporal variation in the dataset; all generators reproduce these distributions accurately. The binary outcome label y shows a strong class imbalance (~89% no, ~11% yes) that is preserved by all generators.

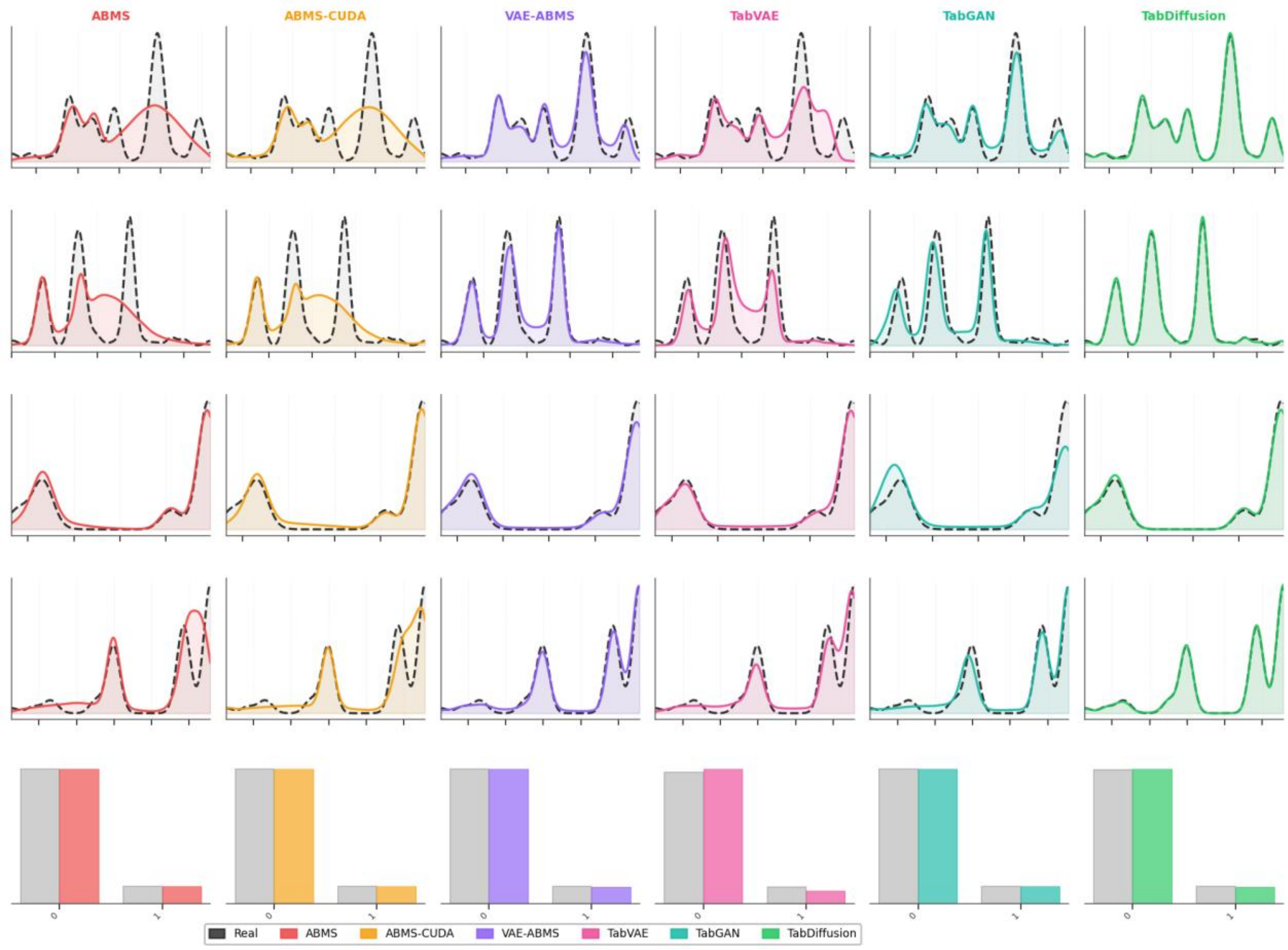

**Supplementary Figure 5.** KDE density overlays for Bank Marketing, features 17–21.

### *8.2.3. NSL-KDD dataset*

Supplementary Fig. 6 covers NSL-KDD features 9–16 (num_failed_logins through num_compromised). Most of these are sparse integer-valued flags or low-count variables with heavy zero-inflation; all generators reproduce the dominant zero mass, though TabDiffusion most accurately captures the small non-zero tail masses.

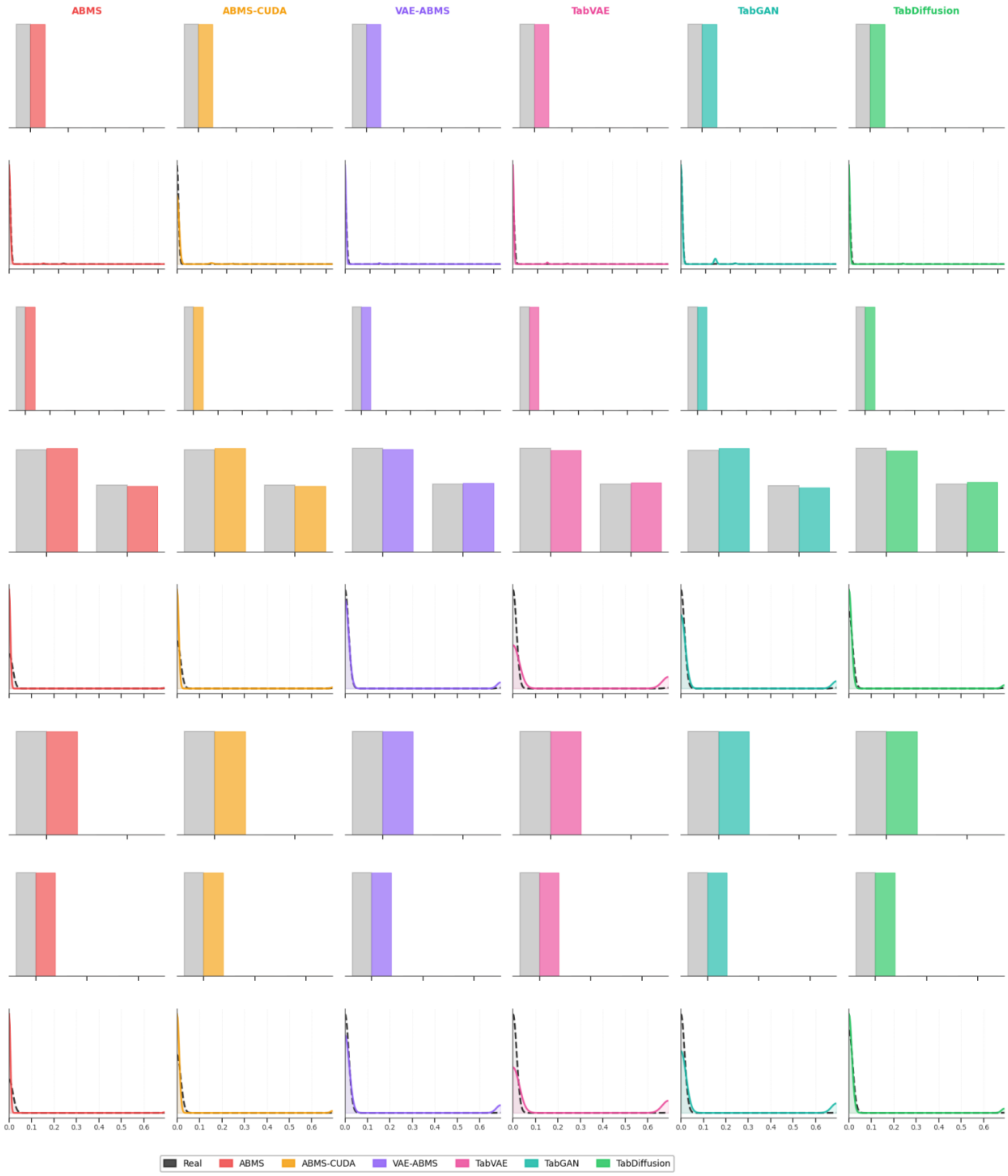

**Supplementary Figure 6.** KDE density overlays for NSL-KDD, features 9–16.

Supplementary Fig. 7 covers features 17–24 (root_shell through dst_bytes). The dst_bytes feature exhibits an extremely heavy right tail spanning several orders of magnitude; the log-scale transformation applied automatically by TDGT's pipeline reveals that TabDiffusion and TabGAN best recover the long-tail structure, while ABMS and TabVAE underestimate the tail.

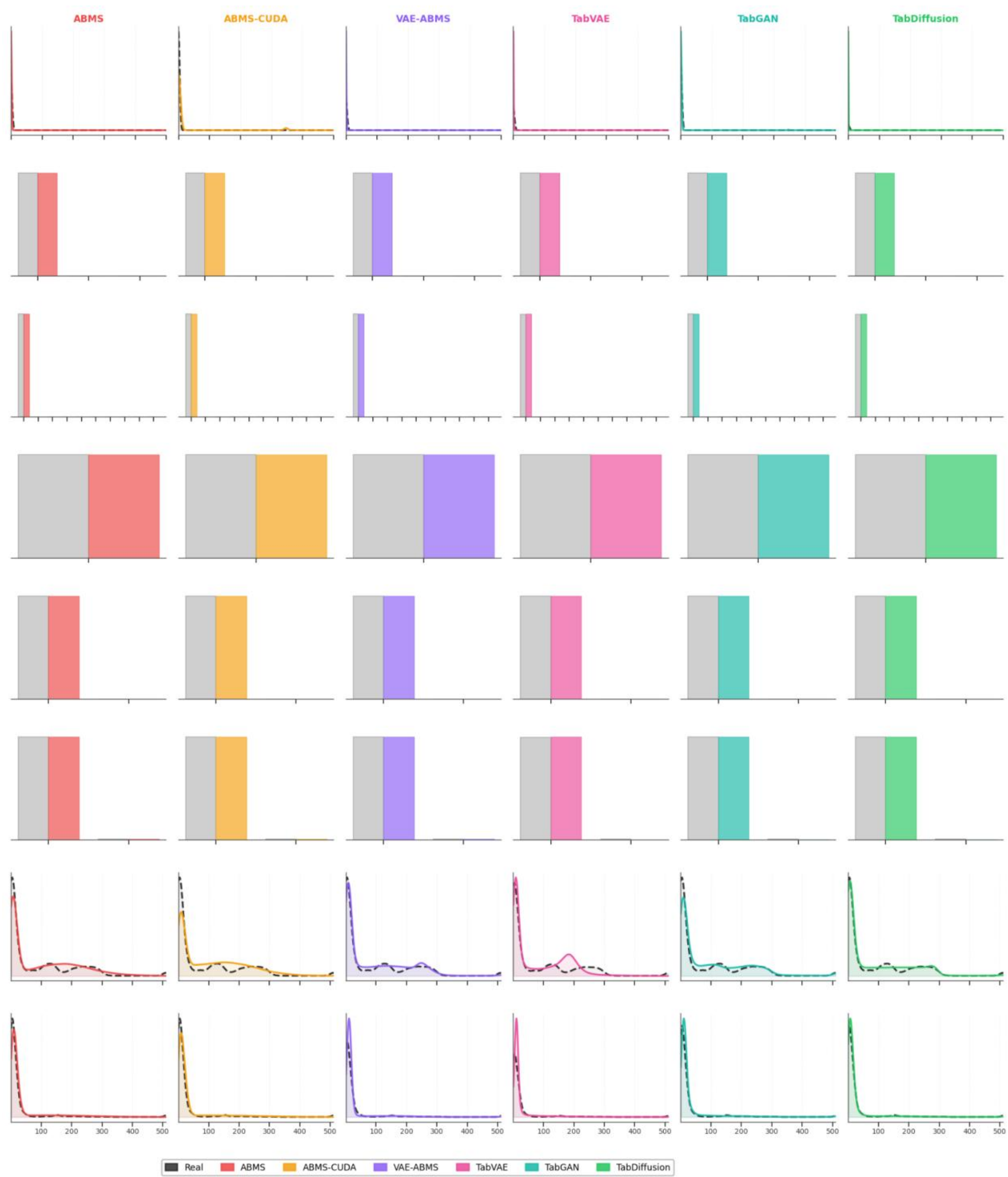

**Supplementary Figure 7.** KDE density overlays for NSL-KDD, features 17–24.

Supplementary Fig. 8 covers features 25–32 (land through dst_host_count). This group includes binary flags (land, urgent) and host-level count variables. Binary flags are shown as bar charts and are reproduced accurately by all generators. Host count variables follow approximately uniform distributions that all generators reproduce well.

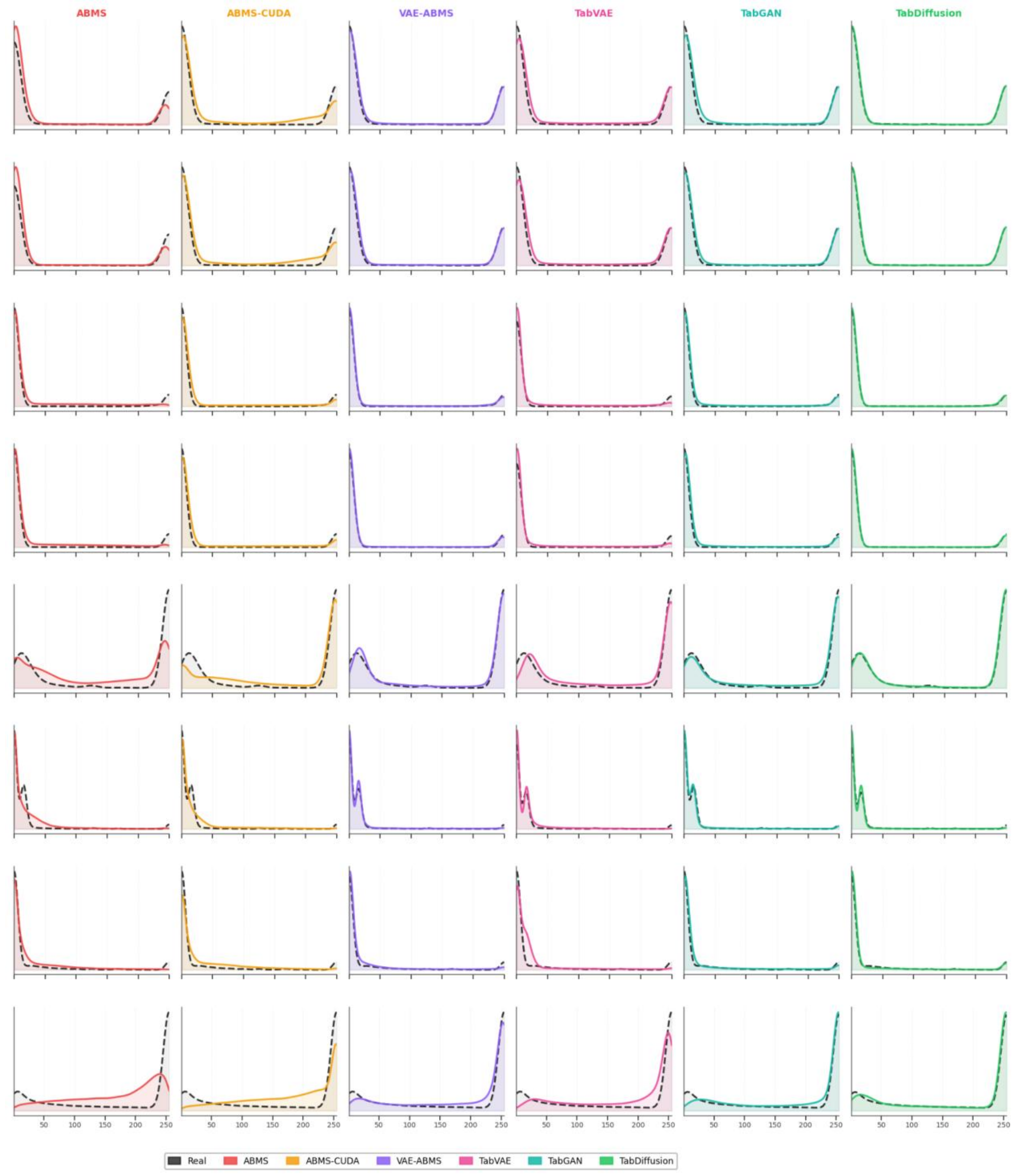

**Supplementary Figure 8.** KDE density overlays for NSL-KDD, features 25–32.

Supplementary Fig. 9 covers features 33–40, including the dst_host_srv_count and related same-service and same-port rate features. These rate features cluster near 0 and 1 with a U-shaped distribution; TabDiffusion captures this bimodality most accurately, while VAE-ABMS tends to produce intermediate values.

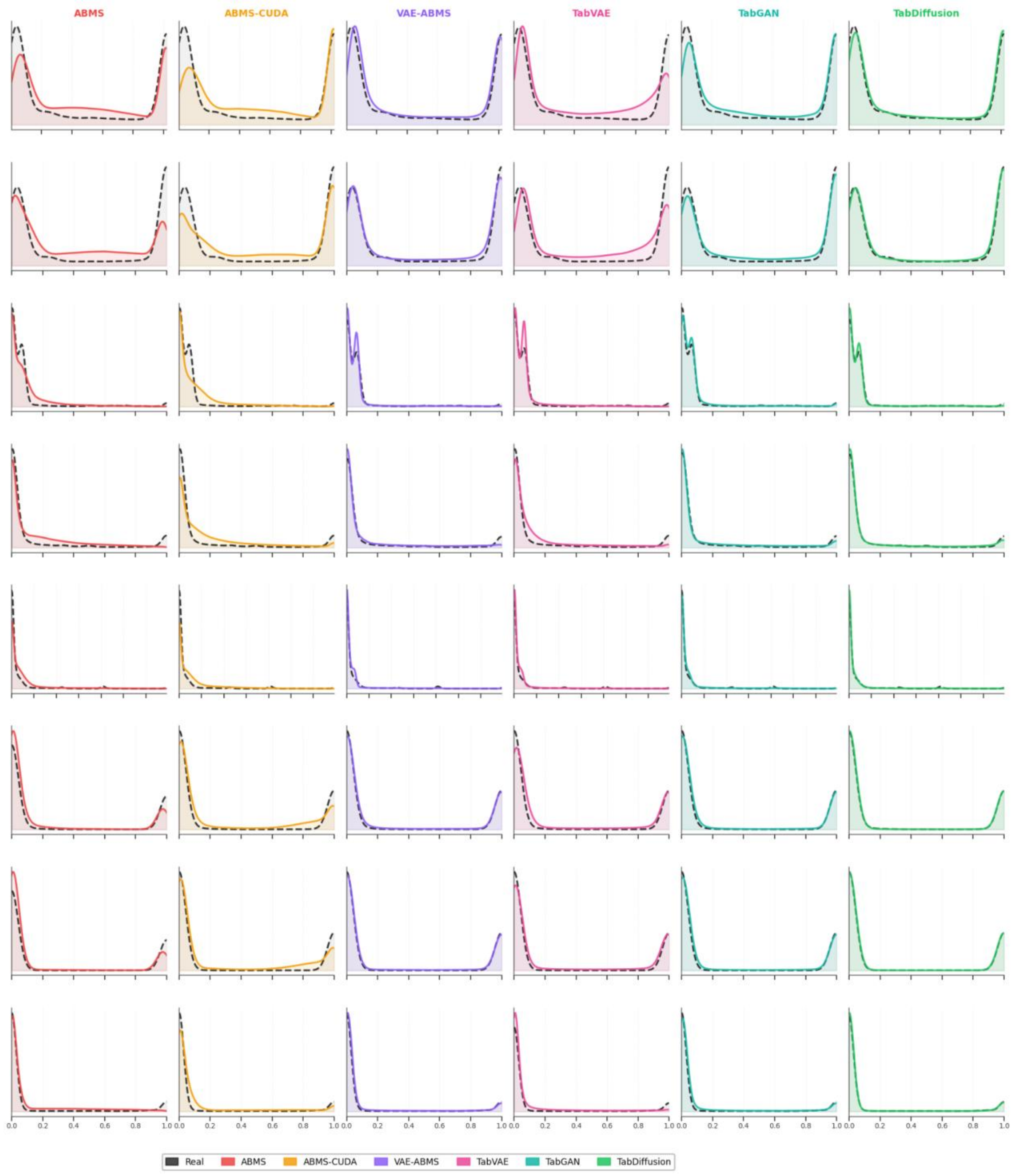

**Supplementary Figure 9.** KDE density overlays for NSL-KDD, features 33–40.

Supplementary Fig. 10 covers the final two NSL-KDD features: dst_host_rerror_rate and the multi-class attack label. The label distribution reflects the significant class imbalance of the dataset (multiple attack subtypes vs. normal traffic); all generators broadly preserve the dominant class proportions, with TabDiffusion and TabGAN showing the closest frequency alignment.

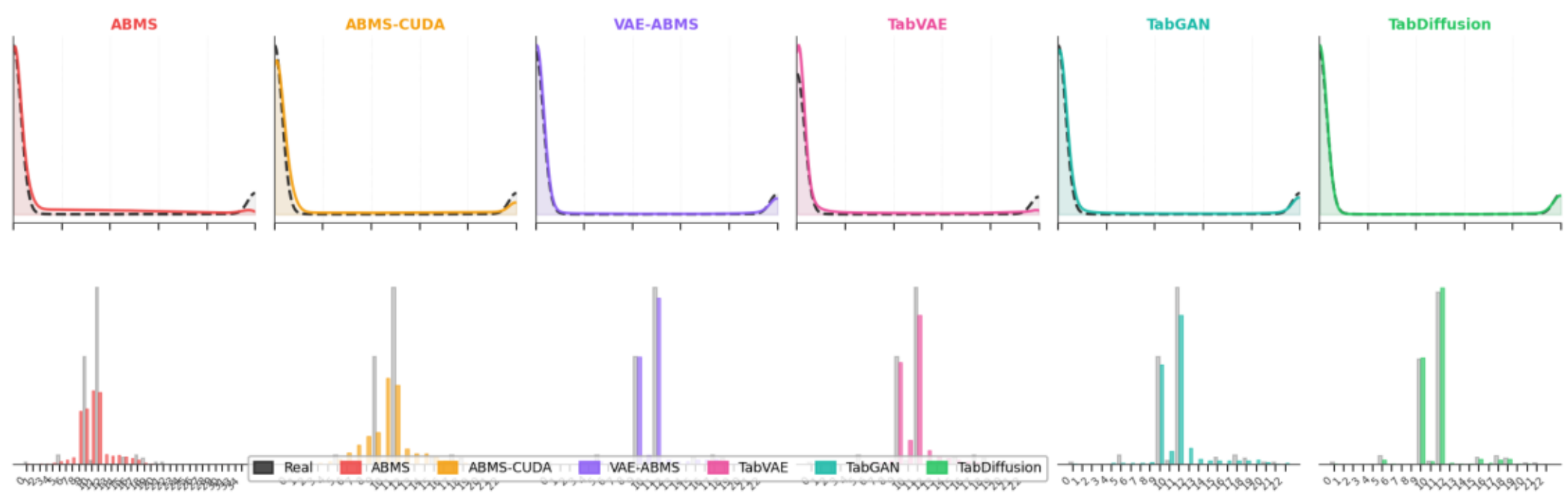

**Supplementary Figure 10.** KDE density overlays for NSL-KDD, features 41–42.

## 8.3. Instances from the TDGT web-based toolkit

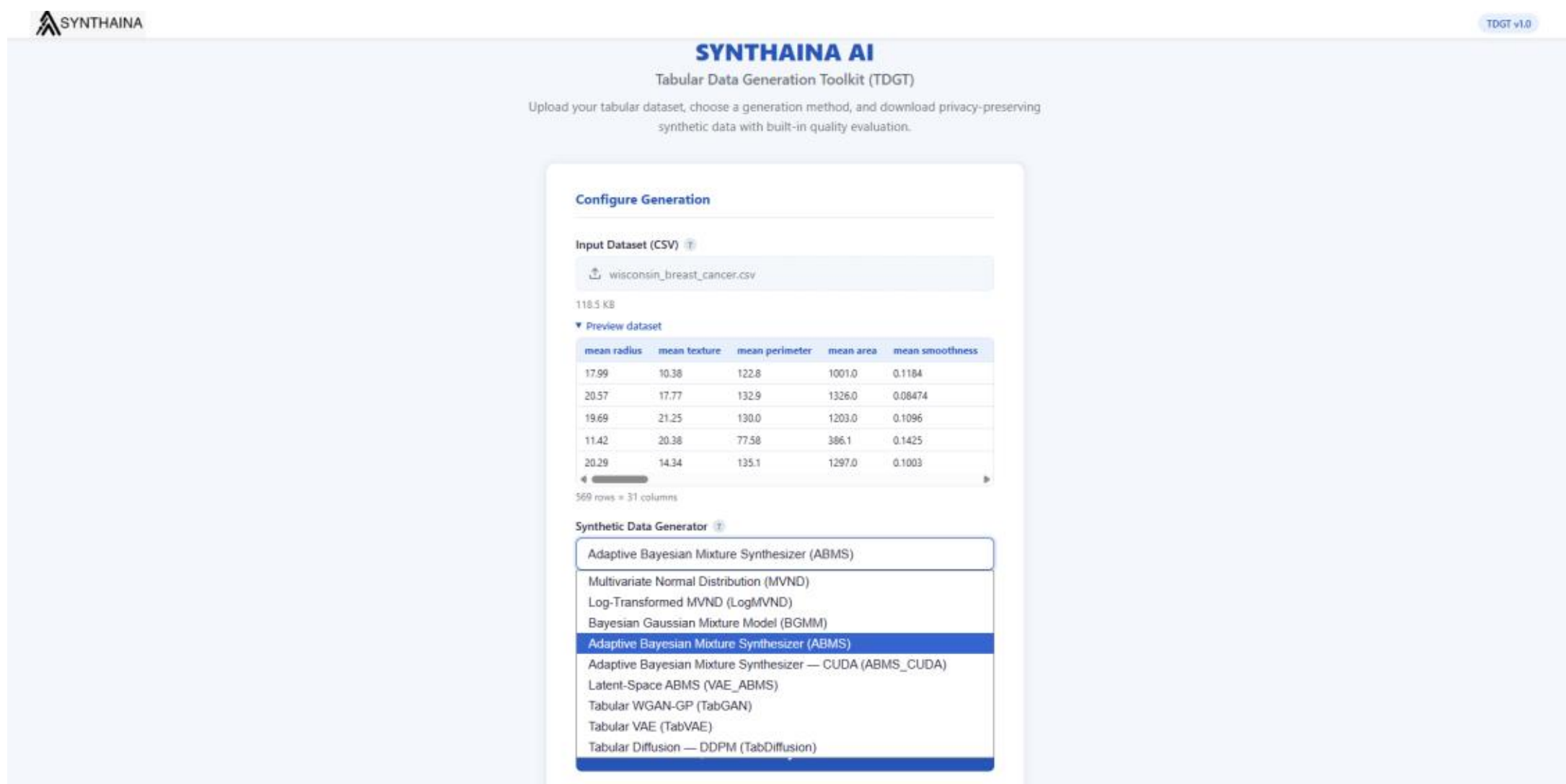

**Supplementary Figure 11.** TDGT data configuration panel — dataset upload and generator selection. The Wisconsin Breast Cancer dataset (wisconsin_breast_cancer.csv, 569 rows × 31 columns) has been uploaded and a preview of the first five rows is rendered automatically, displaying the five leftmost features (mean radius through mean smoothness). The Synthetic Data Generator dropdown is expanded, showing the complete list of available methods.

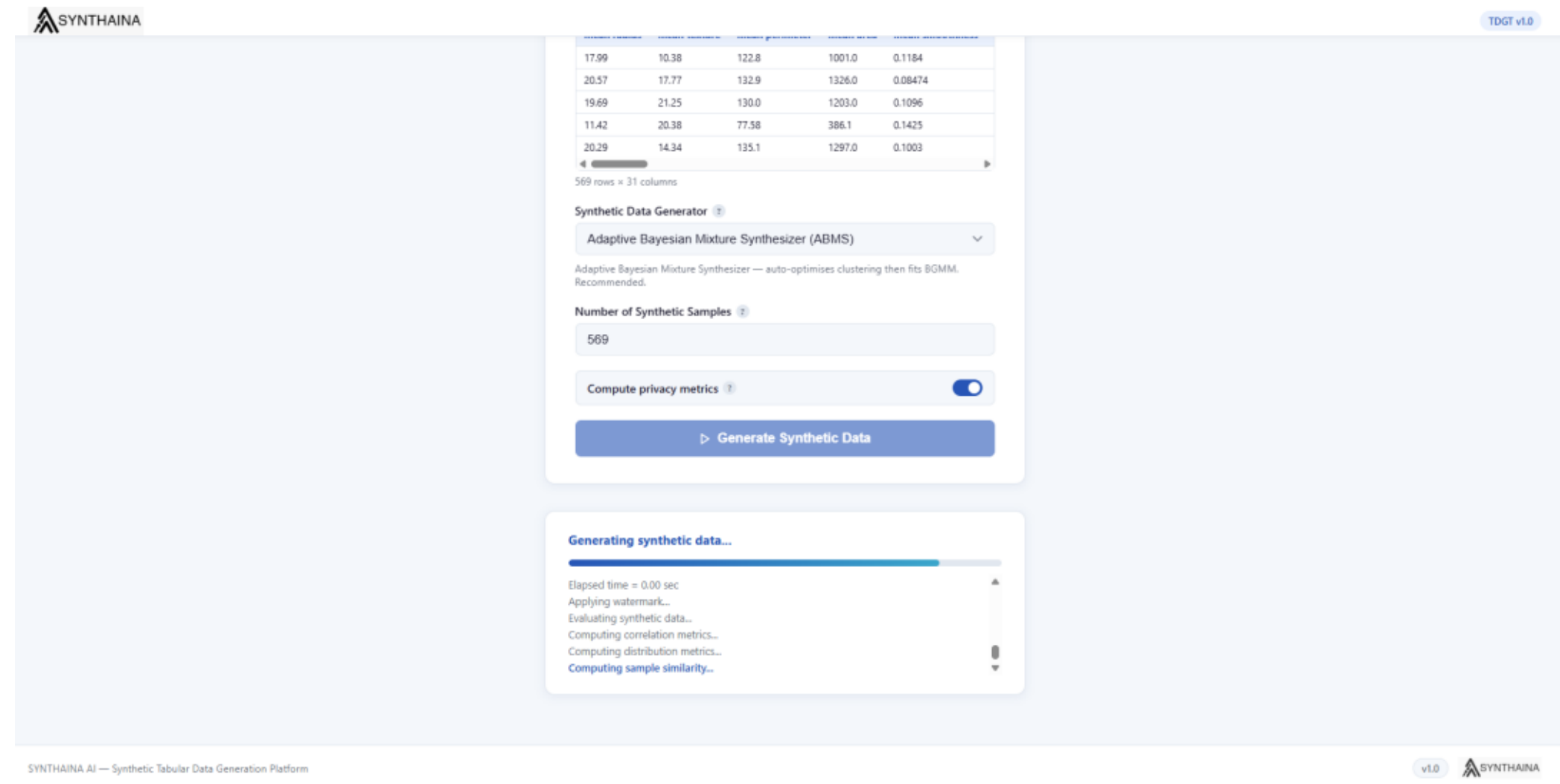

**Supplementary Figure 12.** TDGT real-time generation progress panel. The ABMS generator has been submitted for the Wisconsin Breast Cancer dataset (569 synthetic samples requested, privacy metrics enabled). The animated

progress bar reflects near-complete execution, and the scrollable log reports the sequential pipeline stages.Elapsed time is reported continuously during execution.

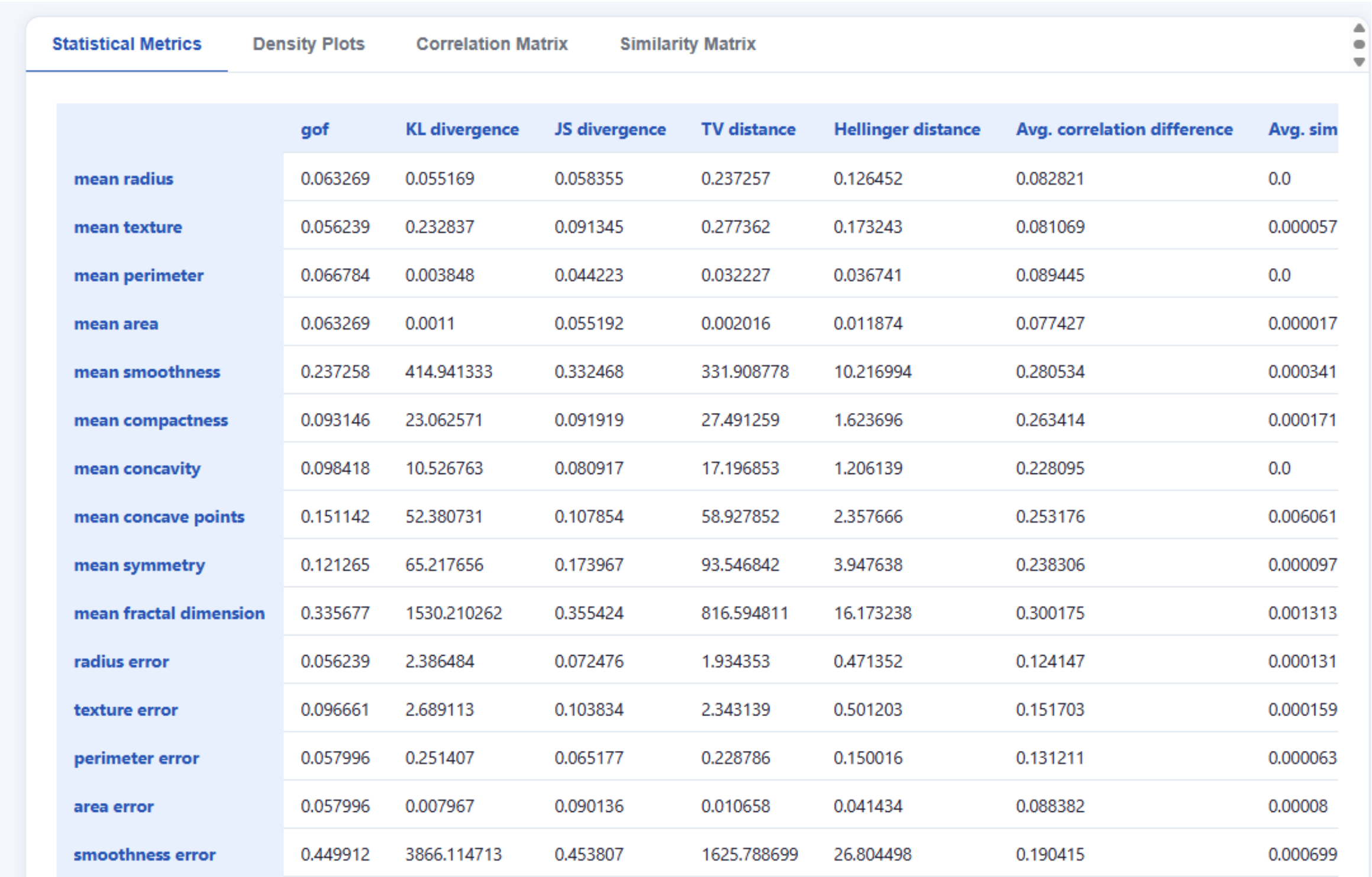

Statistical Metrics | Density Plots | Correlation Matrix | Similarity Matrix

| | gof | KL divergence | JS divergence | TV distance | Hellinger distance | Avg. correlation difference | Avg. sim |
|---|---|---|---|---|---|---|---|
| mean radius | 0.063269 | 0.055169 | 0.058355 | 0.237257 | 0.126452 | 0.082821 | 0.0 |
| mean texture | 0.056239 | 0.232837 | 0.091345 | 0.277362 | 0.173243 | 0.081069 | 0.000057 |
| mean perimeter | 0.066784 | 0.003848 | 0.044223 | 0.032227 | 0.036741 | 0.089445 | 0.0 |
| mean area | 0.063269 | 0.0011 | 0.055192 | 0.002016 | 0.011874 | 0.077427 | 0.000017 |
| mean smoothness | 0.237258 | 414.941333 | 0.332468 | 331.908778 | 10.216994 | 0.280534 | 0.000341 |
| mean compactness | 0.093146 | 23.062571 | 0.091919 | 27.491259 | 1.623696 | 0.263414 | 0.000171 |
| mean concavity | 0.098418 | 10.526763 | 0.080917 | 17.196853 | 1.206139 | 0.228095 | 0.0 |
| mean concave points | 0.151142 | 52.380731 | 0.107854 | 58.927852 | 2.357666 | 0.253176 | 0.006061 |
| mean symmetry | 0.121265 | 65.217656 | 0.173967 | 93.546842 | 3.947638 | 0.238306 | 0.000097 |
| mean fractal dimension | 0.335677 | 1530.210262 | 0.355424 | 816.594811 | 16.173238 | 0.300175 | 0.001313 |
| radius error | 0.056239 | 2.386484 | 0.072476 | 1.934353 | 0.471352 | 0.124147 | 0.000131 |
| texture error | 0.096661 | 2.689113 | 0.103834 | 2.343139 | 0.501203 | 0.151703 | 0.000159 |
| perimeter error | 0.057996 | 0.251407 | 0.065177 | 0.228786 | 0.150016 | 0.131211 | 0.000063 |
| area error | 0.057996 | 0.007967 | 0.090136 | 0.010658 | 0.041434 | 0.088382 | 0.00008 |
| smoothness error | 0.449912 | 3866.114713 | 0.453807 | 1625.788699 | 26.804498 | 0.190415 | 0.000699 |

**Supplementary Figure 13.** TDGT results panel — Statistical Metrics tab (features 1–15). The per-feature evaluation table reports seven metrics for each feature: Goodness-of-Fit (GoF), KL divergence, JS divergence, TV distance, Hellinger distance, average correlation difference (Avg. correlation difference), and average sample similarity (Avg. sim.). Features shown span mean radius through smoothness error.

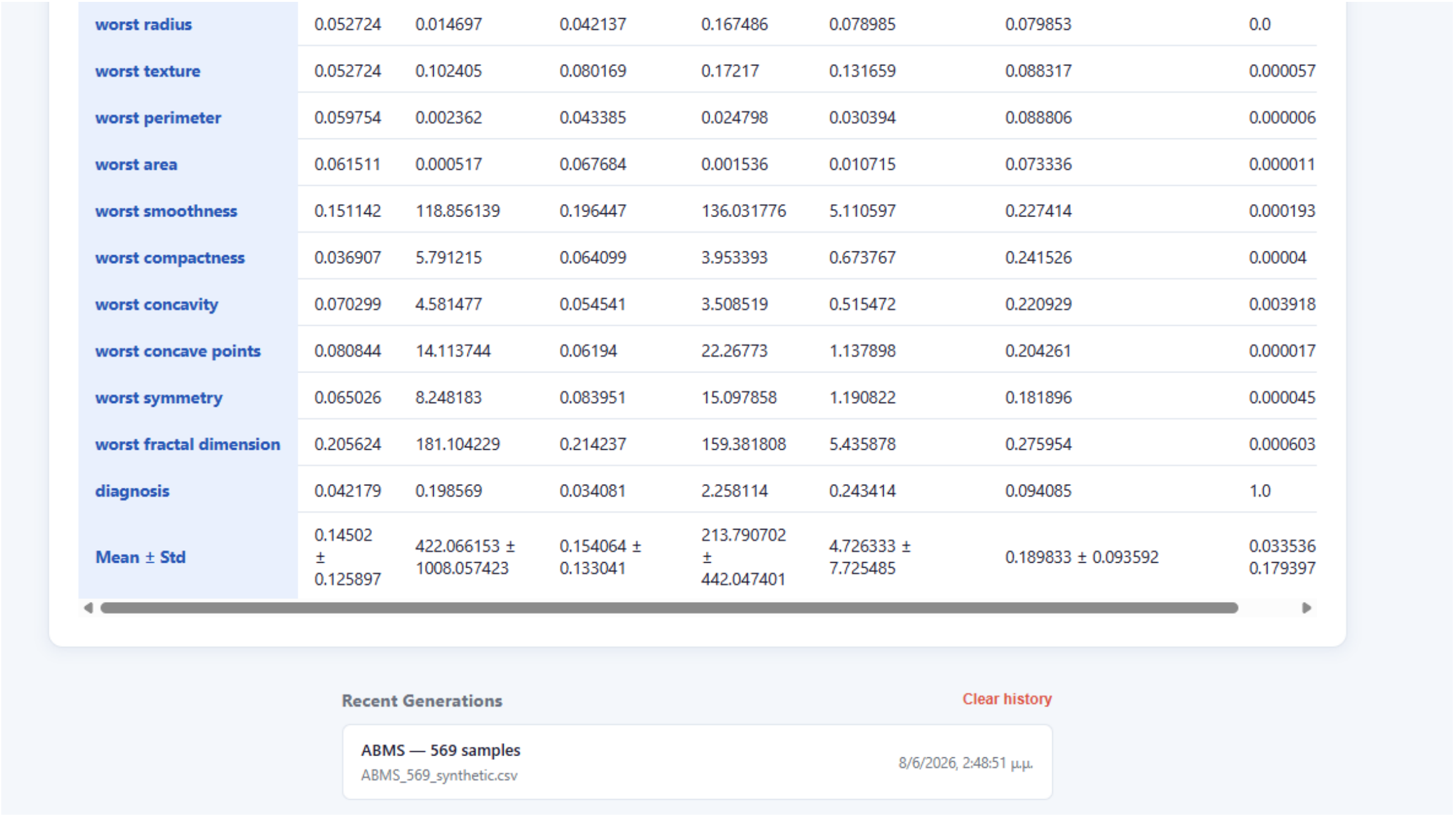

| | | | | | | | |
|---|---|---|---|---|---|---|---|
| worst radius | 0.052724 | 0.014697 | 0.042137 | 0.167486 | 0.078985 | 0.079853 | 0.0 |
| worst texture | 0.052724 | 0.102405 | 0.080169 | 0.17217 | 0.131659 | 0.088317 | 0.000057 |
| worst perimeter | 0.059754 | 0.002362 | 0.043385 | 0.024798 | 0.030394 | 0.088806 | 0.000006 |
| worst area | 0.061511 | 0.000517 | 0.067684 | 0.001536 | 0.010715 | 0.073336 | 0.000011 |
| worst smoothness | 0.151142 | 118.856139 | 0.196447 | 136.031776 | 5.110597 | 0.227414 | 0.000193 |
| worst compactness | 0.036907 | 5.791215 | 0.064099 | 3.953393 | 0.673767 | 0.241526 | 0.00004 |
| worst concavity | 0.070299 | 4.581477 | 0.054541 | 3.508519 | 0.515472 | 0.220929 | 0.003918 |
| worst concave points | 0.080844 | 14.113744 | 0.06194 | 22.26773 | 1.137898 | 0.204261 | 0.000017 |
| worst symmetry | 0.065026 | 8.248183 | 0.083951 | 15.097858 | 1.190822 | 0.181896 | 0.000045 |
| worst fractal dimension | 0.205624 | 181.104229 | 0.214237 | 159.381808 | 5.435878 | 0.275954 | 0.000603 |
| diagnosis | 0.042179 | 0.198569 | 0.034081 | 2.258114 | 0.243414 | 0.094085 | 1.0 |
| Mean ± Std | 0.14502 ± 0.125897 | 422.066153 ± 1008.057423 | 0.154064 ± 0.133041 | 213.790702 ± 442.047401 | 4.726333 ± 7.725485 | 0.189833 ± 0.093592 | 0.033536 0.179397 |

Recent Generations
Clear history
ABMS — 569 samples
ABMS_569_synthetic.csv
8/6/2026, 2:48:51 μ.μ.

**Supplementary Figure 14.** TDGT results panel — Statistical Metrics tab (features 16–31 and summary row). The table continues from Supplementary Fig. 13, covering perimeter error through the binary diagnosis feature, followed by a Mean ± Std summary row reporting the mean and standard deviation of each metric across all 31 features (GoF: 0.145 ± 0.126; KL divergence: 422.07 ± 1008.06; JS divergence: 0.154 ± 0.133; TV distance: 213.79 ± 442.05; Hellinger distance: 4.726 ± 7.725; Avg. correlation difference: 0.190 ± 0.094; Avg. sim.: 0.034

± 0.179). A Recent Generations history entry (ABMS — 569 samples, ABMS_569_synthetic.csv) is visible at the bottom of the interface.

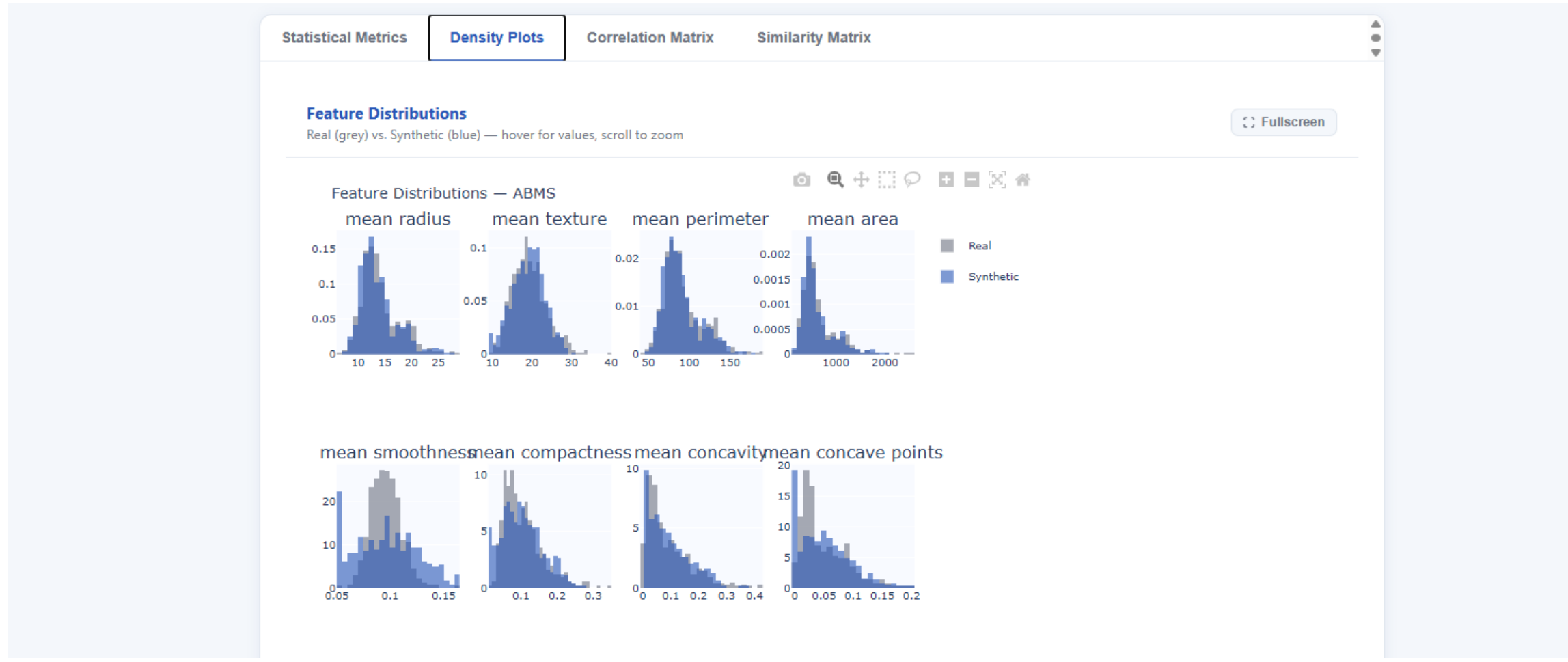

**Supplementary Figure 15.** TDGT results panel — Density Plots tab, features 1–8. The Density Plots tab is selected, revealing the Feature Distributions panel with the subtitle "Real (grey) vs. Synthetic (blue) — hover for values, scroll to zoom." Overlaid histograms are displayed for the first eight features of the Wisconsin Breast Cancer dataset: mean radius, mean texture, mean perimeter, mean area, mean smoothness, mean compactness, mean concavity, and mean concave points. Grey bars represent the real data distribution and blue bars represent the ABMS-generated synthetic distribution. The Fullscreen expansion button is available at the top right of the panel.

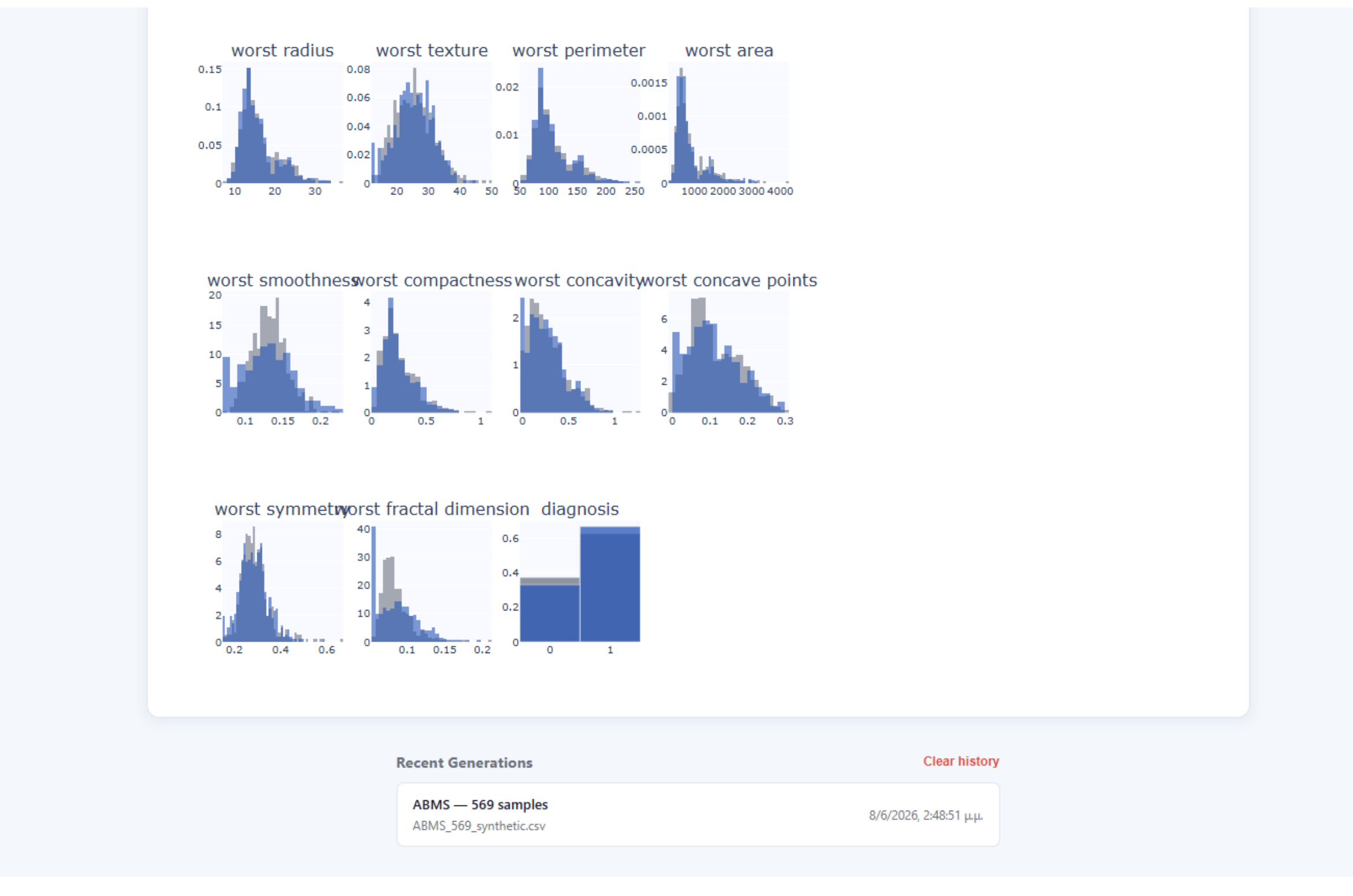

**Supplementary Figure 16.** TDGT results panel — Density Plots tab, features 9–31 (worst measurements and diagnosis). Overlaid histograms continue for the remaining features: worst radius, worst texture, worst perimeter, worst area, worst smoothness, worst compactness, worst concavity, worst concave points, worst symmetry, worst fractal dimension, and the binary diagnosis label. The diagnosis feature is rendered as a grouped bar chart reflecting the class balance of the dataset. The ABMS synthetic distribution (blue) broadly follows the real distribution (grey) across the worst-value features, with moderate deviations on worst fractal dimension and worst concave points.

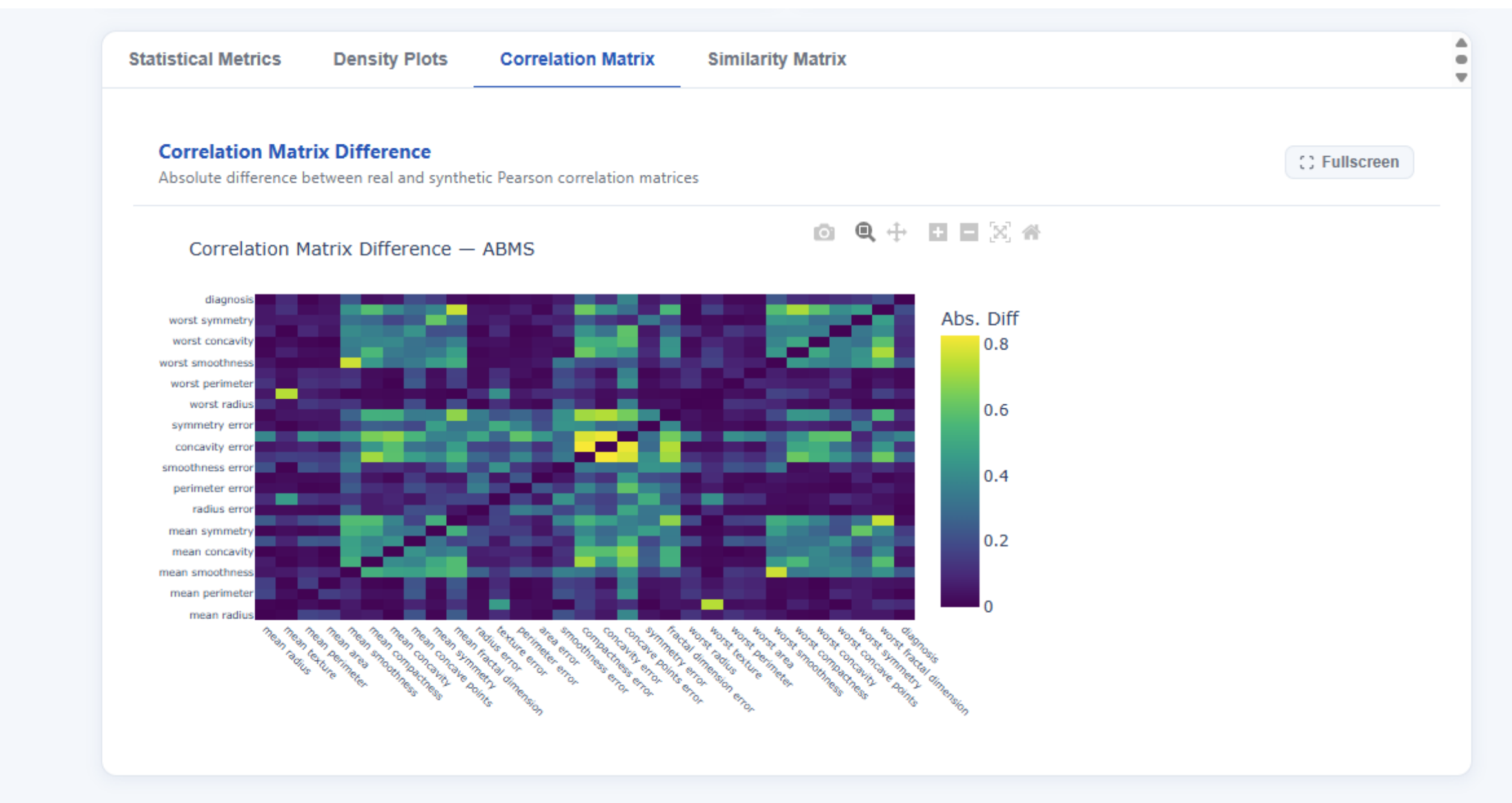

**Supplementary Figure 17.** TDGT results panel — Correlation Matrix tab. The tab displays the Correlation Matrix Difference heatmap, labelled "Absolute difference between real and synthetic Pearson correlation matrices." The 31 × 31 heatmap visualises the absolute element-wise deviation between the real and synthetic Pearson correlation matrices for all feature pairs in the Wisconsin Breast Cancer dataset, using a viridis colour scale (dark purple = negligible difference, yellow = large difference). Notable deviations are visible between the error-scale features (smoothness error, concavity error, symmetry error) and the mean/worst feature blocks, consistent with the ABMS mean correlation matrix difference of ΔC = 0.184 reported in Table 2. The Fullscreen button enables inspection of individual cell values via hover annotation.

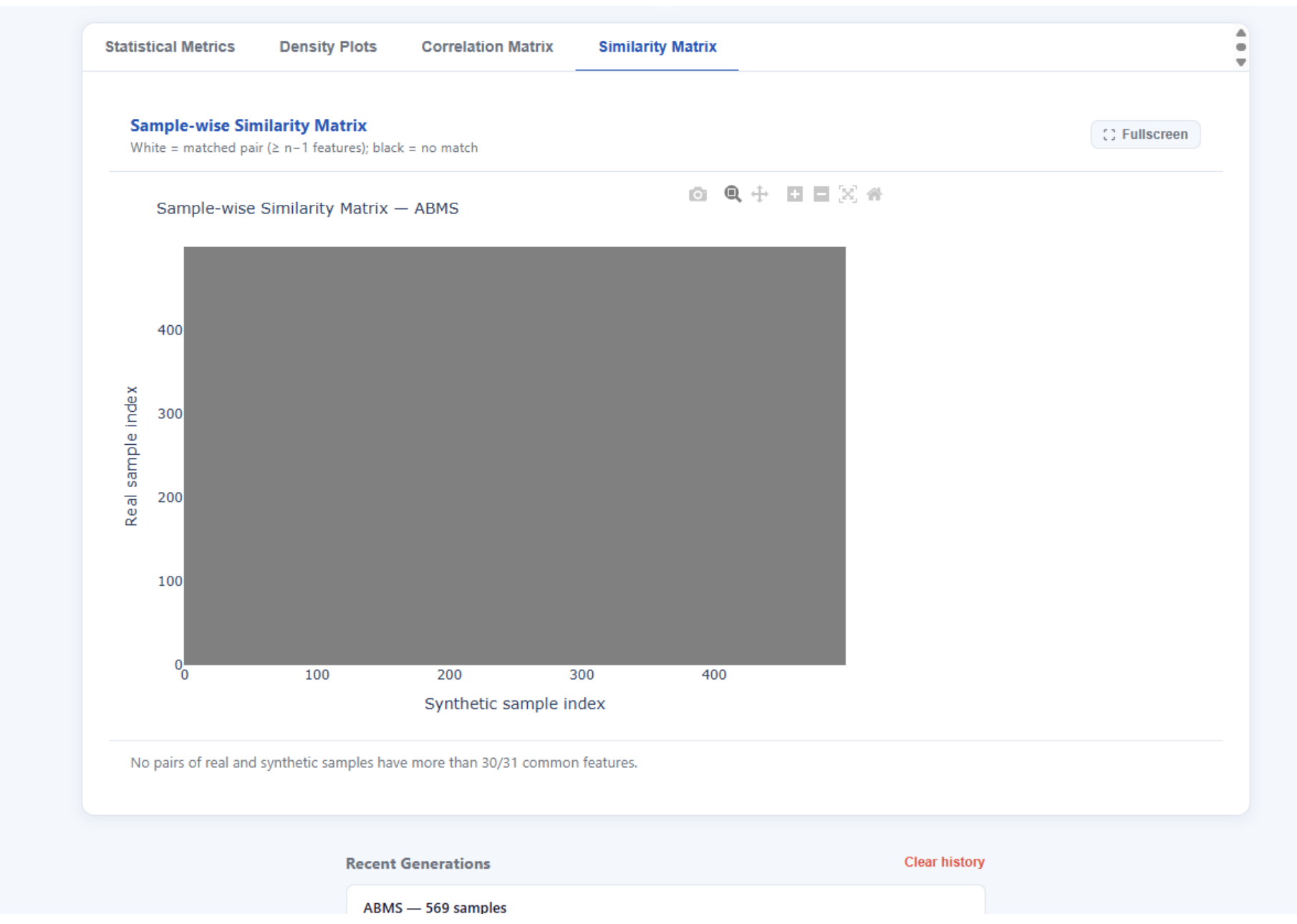

**Supplementary Figure 18.** TDGT results panel — Similarity Matrix tab. The Sample-wise Similarity Matrix is displayed with the subtitle "White = matched pair ($\geq n-1$ features); black = no match." The 569 × 569 binary matrix comparing each real sample (y-axis) against each synthetic sample (x-axis) is uniformly grey, indicating that no pair of real and synthetic samples agrees on at least 30 out of 31 features. This is confirmed by the note below the matrix: "No pairs of real and synthetic samples have more than 30/31 common features," and is consistent with the 0.00% exact-match disclosure risk reported in the Statistical Metrics tab. The Recent Generations history panel at the bottom records the completed ABMS run.